\documentclass[11pt]{article}

\usepackage[final]{acl}

\usepackage{times}
\usepackage{latexsym}

\usepackage[T1]{fontenc}

\usepackage[utf8]{inputenc}

\usepackage{microtype}

\usepackage{inconsolata}

\usepackage{graphicx}

\usepackage{algorithm}
\usepackage{algorithmic}
\usepackage{booktabs}
\usepackage{amsmath}
\usepackage[table]{xcolor} 
\usepackage{tabularx} 
\usepackage{graphicx}    
\usepackage{multirow}    
\usepackage{diagbox}
\usepackage{subcaption}

\newcommand{\emphasize}[1]{``#1''}

\title{PREFINE: Personalized Story Generation via Simulated User Critics and User-Specific Rubric Generation}

\author{
 \textbf{Kentaro Ueda\textsuperscript{1,3}},
 \textbf{Takehiro Takayanagi\textsuperscript{2,3}},
\\
 \textsuperscript{1}Nara Institute of Science and Technology,\\
 \textsuperscript{2}The University of Tokyo,\\
 \textsuperscript{3}Simulacra Inc.\\
 \texttt{k\_ueda@simulacra.co.jp}, \texttt{takayanagi@simulacra.co.jp}
}

\begin{document}
\maketitle
\begin{abstract}
Personalizing story generation to individual users remains a core challenge in natural language generation. Existing approaches typically require explicit user feedback or fine-tuning, which pose practical concerns in terms of usability, scalability, and privacy. In this work, we introduce PREFINE (Persona-and-Rubric Guided Critique-and-Refine), a novel Critique-and-Refine framework that enables personalized story generation without user feedback or parameter updates. PREFINE constructs a pseudo-user agent from a user's interaction history and generates user-specific rubrics (evaluation criteria). These components are used to critique and iteratively refine story drafts toward the user’s preferences. We evaluate PREFINE on two benchmark datasets, PerDOC and PerMPST, and compare it with existing approaches. Both automatic and human evaluations show that PREFINE achieves significantly better personalization while preserving general story quality. Notably, PREFINE outperforms existing in-context personalization and critique-based generation methods, and can even enhance already personalized outputs through post-hoc refinement. Our analysis reveals that user-specific rubrics are critical in driving personalization. The results demonstrate the effectiveness and practicality of inference-only, rubric-guided personalization, with potential applications beyond storytelling, including dialogue, recommendation, and education.
\end{abstract}

\begin{figure}[h]
  \centering
  \includegraphics[width=\columnwidth]{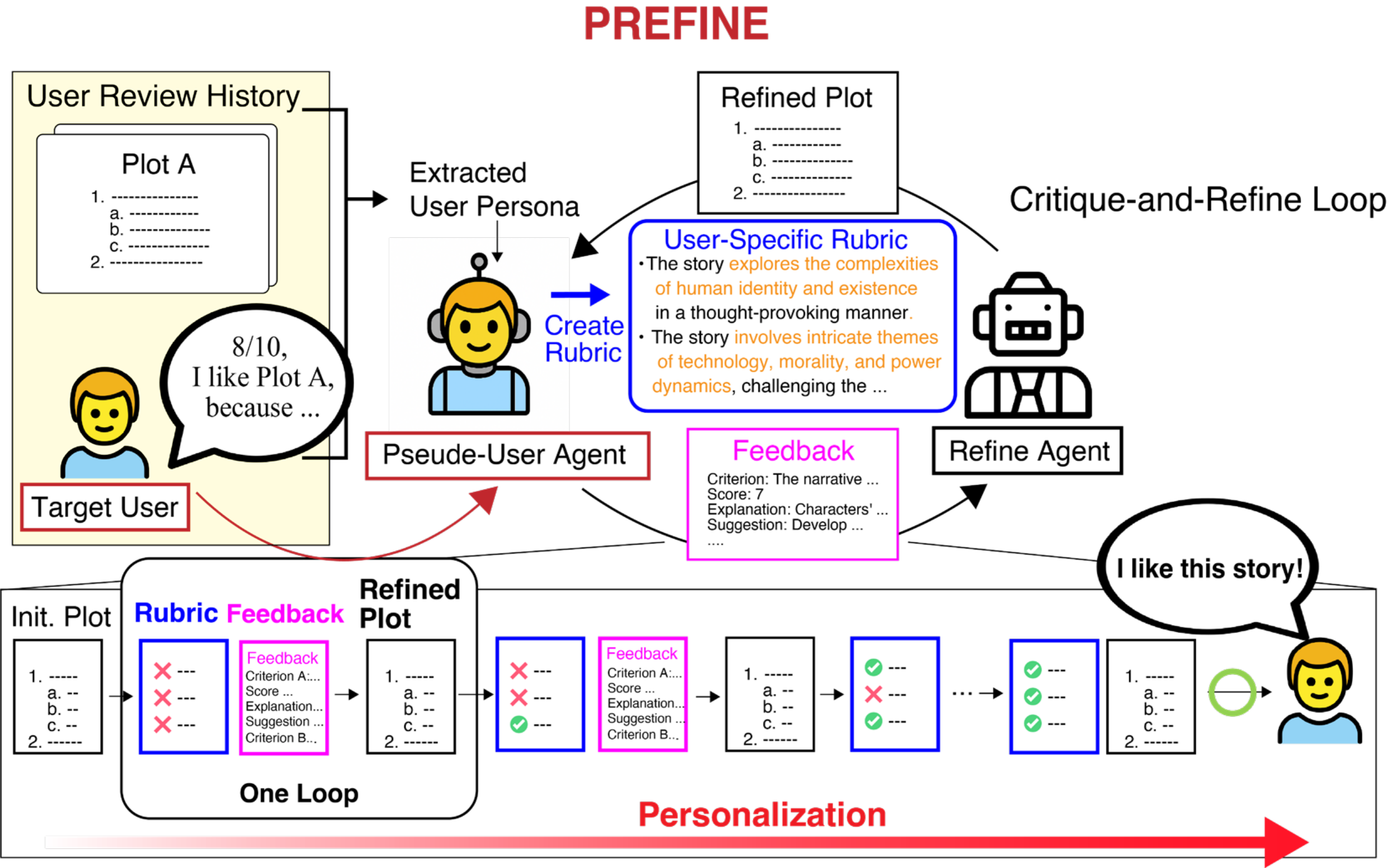}
  \caption{Overview of PREFINE. PREFINE extends the Critique-and-Refine framework by introducing a pseudo-user agent and a user-specific rubric. This enables the generation of personalized stories for the target user without relying on user feedback or additional model fine-tuning.}
  \label{fig:overview}
\end{figure}

\section{Introduction}

Recent advances in Large Language Models (LLMs) have significantly improved performance on creative text generation tasks, such as storytelling and plot synthesis \cite{e2e,yang-etal-2022-re3}. However, most LLMs are primarily optimized for general-purpose output quality, and challenges remain in generating personalized stories that reflect individual user preferences \cite{perse}. User preferences, including favored character types and narrative tones, are highly diverse and idiosyncratic, requiring story generation systems to dynamically adapt to each user.

Traditional personalization methods rely on explicit user feedback or fine-tuning \cite{finetuning-persona}, but they incur user effort, high training cost, scarce data, and privacy risks.

Recently, Critique-and-Refine (C\&R) frameworks, including Self-Refine \cite{self-refine} and CRITIC \cite{conf/iclr/2024}, have garnered attention for improving output quality without relying on human feedback or model retraining. In these approaches, LLMs critique and refine their outputs based on predefined evaluation criteria (rubrics), thereby enhancing quality. 
Although existing C\&R frameworks have proven effective for general quality improvement, they lack mechanisms to incorporate user preferences into the generation process and thus fall short of enabling personalized generation \cite{cr-survey}.

We propose PREFINE (Persona-and-Rubric Guided Critique-and-Refine), a novel C\&R-based framework extended for personalization. PREFINE enables personalized generation without parameter updates or continuous user feedback. PREFINE consists of two key components:

First, we introduce a pseudo-user agent. Prior research has shown that LLMs can effectively mimic user preferences and styles \cite{frompersona, personallm, park2024generativeagentsimulations1000}. Building on this, PREFINE constructs a pseudo-user agent that acts on behalf of the user, enabling the C\&R process to adapt to individual preferences without requiring model retraining or iterative user feedback.

Second, we introduce user-specific rubric generation: unlike fixed, designer-defined rubrics in conventional C\&R, PREFINE derives rubrics from each user’s interaction history to enable more accurate critique and refinement.

We evaluate PREFINE on two story generation datasets, PerDOC \cite{perdoc} and PerMPST \cite{mpst}, which include user interaction histories. 
We compare against representative non-personalized, prompt-based personalization, and a C\&R for general story-quality improvement, and conduct ablations on the pseudo-user and rubric mechanisms.
Both automatic and human evaluations were conducted. For automatic evaluation, we adopt an LLM-as-a-judge framework, with judging quality verified in preliminary experiments.

PREFINE outperforms all baselines and variants in most cases, achieving significantly higher win rates in PerDOC and scores in PerMPST. The ablation analysis confirms the benefit of both the advanced persona modeling and personalized rubrics.

Our contributions: (i) we propose PREFINE, which extends Critique-and-Refine to personalized story generation by constructing a pseudo-user agent from user history and leveraging user-specific rubrics; (ii) PREFINE delivers improved personalization with story quality on par with specialized enhancement methods.

\section{Related Work}
\subsection{Personalized Text Generation}

Recent advances in LLMs have shifted the focus from generating high-quality, generic outputs to producing texts that align more closely with individual user preferences \cite{xu-etal-2025-personalized, mok-etal-2025-exploring, fin_advisor}. Traditional approaches to personalization, such as model fine-tuning \cite{finetuning-persona} or reinforcement learning \cite{whosebote}, can be effective, but they come with substantial costs. These include the burden of collecting user-specific preference data, the computational overhead of retraining models, and increased risks related to user privacy during deployment.
To overcome these challenges, recent work has explored in-context learning as a lightweight and privacy-preserving alternative for effective personalization. By conditioning model behavior directly through prompts that embed user preference information, in-context personalization eliminates the need for model updates while adapting to individual users \cite{deshpande-etal-2023-toxicity}.

\subsection{Story Generation}
One domain where personalization is particularly valuable is story generation \cite{peng-etal-2018-towards}. In this task, user preferences such as character traits, genre, and narrative development play a central role. However, most prior research has focused on improving general narrative quality rather than adapting to individual tastes \cite{perse}. For instance, existing methods often rely on structured plot schemas \cite{e2e} or rewriting modules to enhance coherence and resolve contradictions \cite{yang-etal-2022-re3}. \citet{yunusov-etal-2024-mirrorstories} demonstrate that prompting LLMs with readers’ identity attributes (e.g., name and age) enables personalized story generation and leads to increased reader engagement.

More recently, Critique-and-Refine frameworks have gained attention as a lightweight approach for improving story generation without relying on fine-tuning \cite{bae-kim-2024-collective, self-refine, conf/iclr/2024, cr-survey, criticbench}. In a typical C\&R workflow, a language model provides feedback on an initial output and iteratively refines it based on that feedback. Despite their success in general text improvement, existing C\&R methods largely overlook the challenge of personalization \cite{cr-survey}.

\subsection{LLM-based User Modeling}
Recent work has shown that LLMs are capable of simulating user preferences and linguistic styles through prompting \cite{frompersona}. For example, LLMs can partially reproduce Big Five personality traits when appropriately prompted \cite{personallm}, and reflect opinion distributions of specific demographic groups when given relevant demographic cues \cite{whoseopinion}. Park et al.~\cite{park2024generativeagentsimulations1000} further demonstrated that LLMs equipped with expert personas (e.g., psychologists, sociologists) can serve as effective agents for interpreting human behavior and generating high-level insights. In the context of personalized story generation, PerSE \cite{perse} introduces a benchmark and a preference-aware LLM evaluator that measures alignment with user preferences, showing that LLMs prompted with user personas can reliably serve as automatic judges.
Building on these findings, this paper investigates whether pseudo-user agents, constructed using LLMs’ user-simulation capabilities, can serve as critique agents in the C\&R framework to improve personalized story generation.

\section{Method}
\textbf{Problem Formulation}
Formally, personalized story generation for a user $u$ is defined as the task of generating a story $s$ from a given premise $c$ (such as \emphasize{Architect Mark Jacobs returns to Metro City to dedicate the Onyx Skyscraper, which he designed}) such that $s$ aligns as closely as possible with the user's preference $P_u$.
Since $P_u$ is not directly observable, we predict an estimate $\hat{P_u}$ from the user's historical interaction data $H_u$. The format of $H_u$ varies depending on the dataset, consisting of either story-level ratings and comments or pairwise preference annotations between story variants (see Section~\ref{sec:datasets} for details). In all cases, we use $\hat{P_u}$ to guide the generation process.

\subsection{PREFINE}
We propose \textbf{PREFINE} (Persona-and-Rubric guided Critique-and-Refine), a framework for personalized story generation without parameter updates or explicit user feedback. As illustrated in Figure~\ref{fig:overview}, PREFINE consists of three core components: (1) a pseudo-user critique agent that imitates user preferences, (2) user-specific rubric generation tailored to individual users, and (3) a Critique-and-Refine loop that improves story outputs accordingly.
Details of the prompt configurations are provided in Appendix~\ref{appendix:promptsetting}.
Each component of PREFINE is described in the following.

\subsection{Initial Story Generation}
Given a premise $c$, we generate an initial story $s^{(0)}$ using a large language model $\mathcal{M}$. The generation is based on an initial prompt ($\mathsf{prompt}_{\text{init}}$) as follows:

\begin{equation}
s^{(0)} = \mathcal{M}(\mathsf{prompt}_{\text{init}}, c)
\end{equation}

In our framework, PREFINE begins with this initial output $s^{(0)}$ and iteratively improves its alignment with the target user's preferences. The initial story is generated without conditioning on user preferences, in order to establish a user-independent baseline and to enable quantitative evaluation of the personalization effects introduced by PREFINE.

\subsection{Pseudo-User Agent}
The pseudo-user agent \footnote{For notational convenience, we denote by $\mathcal{M}_u$ the model used to simulate user preferences. In practice, $\mathcal{M}_u$ is not a separately trained model, but the base LLM conditioned by a prompt that incorporates the persona description.} $\mathcal{M}_u$ is designed to simulate user $u$ based on their predicted preference $\hat{P}_u$. To construct a faithful pseudo-user agent, we prompt an expert agent $\mathcal{M}_\text{expert}$ to estimate $\hat{P}_u$ from the user's interaction history $H_u$, inspired by prior work simulating domain experts via prompt-based LLMs \cite{park2024generativeagentsimulations1000} (see Appendix~\ref{appnedix:ExpertLLMAgents} for details). The result is expressed as a natural language Explicit Persona (EP), which serves as a surrogate representation of $P_u$ in downstream components.

Acting as a stand-in for the actual user, $\mathcal{M}_u$ is responsible for generating user-specific rubrics and providing critiques and feedback in accordance with the predicted preferences.

\subsection{User-Specific Rubric Generation}
In this step, the pseudo-user agent $\mathcal{M}_u$ transforms the user's preference into a structured rubric that serves as a consistent evaluation standard throughout the refinement cycle. Unlike generic rubrics that focus on surface-level quality metrics such as grammar or coherence, the rubrics generated here are tailored to the preferences of individual users.

Formally, the user-specific rubric $R_u$ is generated as follows:
\begin{equation}
R_u = \mathcal{M}_u(\mathsf{prompt}_{\text{rubric}})
\end{equation}
Here, $\mathcal{M}_u$ denotes the base LLM conditioned on EP via prompt design. The $\mathsf{prompt}_{\text{rubric}}$ specifies an instruction to generate 3–5 evaluation criteria reflecting the user's preference.

The resulting rubric guides both evaluation and refinement within the Critique-and-Refine cycle. It provides consistent, user-aligned feedback at each step, and ensures that improvements remain focused on the target user's preferences and enabling stable and effective personalization.

\subsection{Critique-and-Refine Cycle}
PREFINE employs an iterative Critique-and-Refine cycle to progressively adapt a story to user-specific preferences. In this cycle, a pseudo-user agent provides feedback on the current story, followed by a refinement step that updates the story accordingly.

\paragraph{Critique and Feedback Generation}
At iteration step $t$, the pseudo-user agent $\mathcal{M}_u$ generates feedback $F^{(t)}$ for the current story $s^{(t)}$ based on the user-specific rubric $R_u$:
\begin{equation}
F^{(t)} = \mathcal{M}_u(\mathsf{prompt}_{\text{feedback}}, R_u, s^{(t)})
\end{equation}
where $\mathsf{prompt}_{\text{feedback}}$ instructs the model to provide scores, justifications, and concrete revision suggestions aligned with the criteria in $R_u$. This enables the generation of structured feedback tailored to the user's preferences.

\paragraph{Story Refinement}
Given the feedback $F^{(t)}$, the story refinement agent $\mathcal{M}_{\text{refine}}$ generates the revised story $s^{(t+1)}$:
\begin{equation}
s^{(t+1)} = \mathcal{M}_{\text{refine}}(\mathsf{prompt}_{\text{refine}}, s^{(t)}, F^{(t)})
\end{equation}
where $\mathsf{prompt}_{\text{refine}}$ guides the model to revise the story based on the provided feedback. The refinement agent is expected to incorporate the suggestions while preserving the narrative coherence and stylistic consistency of the story.

\paragraph{Iteration and Convergence}
The Critique-and-Refine cycle is repeated up to $T$ times (with $T \leq 7$ in our experiments), and the final output $s^{(T)}$ is considered the personalized result.

\section{Experimental Settings}
\subsection{Datasets}
\label{sec:datasets}
We conduct experiments on two story generation datasets with user interaction histories: \textbf{PerDOC}~\cite{perdoc} and \textbf{PerMPST}~\cite{mpst}. Our generation and evaluation settings largely follow the prior work~\cite{perse}.

\paragraph{PerDOC}
PerDOC is a story generation dataset in the OpenPlot format that contains pairwise user preference data annotated with explicit evaluation criteria (e.g., Interestingness, Surprise, Adaptability, Character Quality, Ending Satisfaction) \cite{perdoc}. Each user $u$ is associated with a preference history $H_u = { (\text{PlotA}_i, \text{PlotB}_i, \text{choice}i) }_{i=1}^{N_u}$, where each choice reflects the user's preference between two story variants with respect to a given criterion. Due to the same context length limitations as in prior work, we set $N_u = 1$.

In our experiments, we personalize story generation along the specified criterion (e.g., interestingness) to generate stories that are rated more highly along that dimension.

\paragraph{PerMPST}
PerMPST is a dataset constructed from IMDb\footnote{\url{https://www.imdb.com/}} movie reviews. For each user $u$, we construct an interaction history $H_u$ consisting of $K$ interactions:
\begin{equation}
H_u = \{ c_i \}_{i=1}^{K}, \quad c_i = (\text{synopsis}_i, \text{review}_i, \text{score}_i),
\end{equation}
where each $c_i$ contains a movie synopsis, the user’s review, and a 1–10 rating (10 = highest preference).

In our experiments, we set $K = 4$, i.e., each user interaction history includes four interactions, following the observation in the prior work \cite{perse} that this setting yields stable personalization performance.

The main differences between PerDOC and PerMPST lie in plot length and evaluation format. PerMPST contains relatively short plots accompanied by scalar ratings, whereas PerDOC includes longer plots with pairwise preference annotations (see the Appendix~\ref{appendix:token_dist} for details).
Note also that in both datasets, the \textit{premise} used for story generation is not part of the user’s interaction history $H_u$ but is independently extracted.
In total, we collected 955 \textit{premise}–$H_u$ pairs for PerDOC and 900 pairs for PerMPST.

\subsection{Evaluation Method}
\label{sec:evaluation}
To assess how well the generated stories align with user preferences, we adopt a multifaceted evaluation approach combining automatic evaluation using LLMs and human evaluation.

\paragraph{Automatic Evaluation}
Prior work on this dataset has reported that LLM-based story preference judgments show a moderate correlation with human preference annotations~\cite{perse}.
Preliminary validation (Appendix~\ref{appendix:evaluator_selection}) showed reasonable agreement between human labels and LLM-based evaluations, leading us to select the most aligned evaluators: \texttt{PerSE–LLaMA3–8B}~\cite{perse} for PerDOC and \texttt{GPT-4o}\footnote{gpt-4o-2024-08-06} for PerMPST.

Since \texttt{PerSE–LLaMA3–8B} belongs to the same model family as the story generation model used in this study, we examined this configuration to address possible concerns of family-specific bias \cite{llm-fam-bias} by comparing it with \texttt{PerSE–Gemma3}. The results indicated that such family bias had only a limited effect on the evaluation outcomes (see  Appendix~\ref{appendix:evaluator_selection}). For PerDOC, to control for position bias, we judge each pair twice with reversed order, retaining only consistently selected plots as valid votes when computing win rates \cite{gu2025surveyllmasajudge}. For PerMPST, the evaluator assigns an integer score between 1 and 10 to each generated story.

\paragraph{Human Evaluation}
\label{sec:human_evaluation}
To further verify the alignment of generated stories with user preferences, we also conducted a human evaluation. The annotators consisted of $14$ graduate students (master’s or PhD) recruited specifically for this study.

Each annotator first submitted their preference score for stories in 10-point Likert scale and gave a brief comment explaining the reason. Then, given the same premise, they were shown stories generated by different methods and asked to rate how well each story matched their preferences on a 10-point Likert scale. To break ties, they also ranked the stories in each set from most to least preferred. Additionally, annotators rated the suitability of the generated user-specific rubrics using a 5-point Likert scale. The evaluation was conducted end to end via a dedicated web interface.

Each annotator evaluated four sets; each set contained three stories generated from the same premise by three different methods (four distinct premises in total). Further details on the human evaluation design and interface are provided in Appendix~\ref{appendix:human_eval}.

\paragraph{Evaluation Details}
To evaluate the effectiveness of our proposed method, PREFINE, we compare it against three representative baselines and three variants that isolate specific components of the full system.
We compare PREFINE against the following representative baselines.
\begin{description}
    \item[Zero-Persona (ZP)] 
    This method generates stories based solely on the given premise, without any user-specific information. It aims to produce generally high-quality outputs regardless of individual preferences.

    \item[Prompt–Persona (PP)]~\cite{pp-baseline} The user history $H_u$ is directly appended to the prompt in order to implicitly guide the LLM toward the user's preferences. No explicit persona representation or structuring is used.
    
    \item[Self–Refine (SR)]~\cite{madaan2023self} A representative Critique-and-Refine method using a static rubric. We use the same number of refinement steps as in our method, and adopt a fixed rubric based on narrative quality criteria proposed in \cite{hanna}, which include Relevance, Coherence, Empathy, Surprise, Engagement, and Complexity.
\end{description}

\definecolor{myred}{RGB}{255,120,120}
\definecolor{myblue}{RGB}{120,120,255}

\begin{table}[t]
\small
\centering
\begin{tabularx}{\columnwidth}{l|XXXXXXX}
\toprule
\diagbox{A}{B} & ZP & PP & SR & IPIR & EPIR & IPER & EPER \\
\midrule
\midrule
ZP          & -- & \cellcolor{myblue!80} 0.20 & \cellcolor{myblue!96} 0.04 & \cellcolor{myblue!95} 0.05 & \cellcolor{myblue!97} 0.03 & \cellcolor{myblue!99} 0.01 & \cellcolor{myblue!98} 0.02 \\
PP          & \cellcolor{myred!80} 0.80 & -- & \cellcolor{myred!59} 0.59 & \cellcolor{myred!59} 0.59 & \cellcolor{myblue!55} 0.45 & \cellcolor{myblue!66} 0.34 & \cellcolor{myblue!67} 0.33 \\
SR & \cellcolor{myred!96} 0.96 & \cellcolor{myblue!59} 0.41 & -- & \cellcolor{myred!52} 0.52 & \cellcolor{myblue!69} 0.31 & \cellcolor{myblue!82} 0.18 & \cellcolor{myblue!83} 0.17 \\
\midrule
IPIR        & \cellcolor{myred!95} 0.95 & \cellcolor{myblue!59} 0.41 & \cellcolor{myblue!52} 0.48 & -- & \cellcolor{myblue!67} 0.33 & \cellcolor{myblue!86} 0.14 & \cellcolor{myblue!81} 0.19 \\
EPIR        & \cellcolor{myred!97} 0.97 & \cellcolor{myred!55} 0.55 & \cellcolor{myred!69} 0.69 & \cellcolor{myred!67} 0.67 & -- & \cellcolor{myblue!65} 0.35 & \cellcolor{myblue!69} 0.31 \\
IPER        & \cellcolor{myred!99} 0.99 & \cellcolor{myred!66} 0.66 & \cellcolor{myred!82} 0.82 & \cellcolor{myred!86} 0.86 & \cellcolor{myred!65} 0.65 & -- & \cellcolor{myblue!51} 0.49 \\
\midrule
\textbf{EPER}        & \cellcolor{myred!98} 0.98 & \cellcolor{myred!67} 0.67 & \cellcolor{myred!83} 0.83 & \cellcolor{myred!81} 0.81 & \cellcolor{myred!69} 0.69 & \cellcolor{myred!51} 0.51 & -- \\
\bottomrule
\end{tabularx}
\caption{\textbf{Win rate of A side model vs. B side model}, averaged over five perspectives. 
    Red/blue indicates the row/column method is preferred. 
    See Appendix~\ref{appendix:aspect-winrate-perdoc} for per-perspective results.
    SR: Self-Refine~\cite{self-refine}.}
\label{tab:comparison_winrates}
\end{table}

We define three variants of the full \textsc{PREFINE} configuration (denoted as \textbf{EPER}), to investigate the contributions of two key components:  
(i) the use of an Explicit Persona (EP), and  
(ii) the generation and use of a user-specific Explicit Rubric (ER).

\begin{description}
  \item[IPIR] \textit{Implicit Persona, Implicit Rubric.} The explicit persona is removed. Instead, the user history $H_u$ is directly fed into the pseudo-user agent. No user-specific rubric is generated or applied.

  \item[IPER] \textit{Implicit Persona, Explicit Rubric.} The user history $H_u$ is provided instead of an explicit persona, but a user-specific rubric is still generated and applied.

  \item[EPIR] \textit{Explicit Persona, Implicit Rubric.} An explicit persona (EP) is used to simulate the user, but no user-specific rubric is generated or applied.
\end{description}

For model configuration, we use \texttt{LLaMA-3-70B}\footnote{meta-llama/Llama-3.3-70B-Instruct} as the backbone model for all agent roles (generation, expert, pseudo-user, and refinement), with role-specific prompt designs.
To examine generalizability, we additionally conduct experiments using \texttt{Mistral-7B}\footnote{mistralai/Mistral-7B-Instruct-v0.3} as an alternative backbone.
Implementation details are provided in Appendix~\ref{appendix:Implementation}.

\section{Results}
\subsection{Personalization Results on PerDOC}
\label{sec:perdoc}

\begin{table}[t]
\centering
\footnotesize
\setlength{\tabcolsep}{4pt} 
\begin{tabular}{lccccc}
\toprule
Method              & Score & $\Delta$ & 95\% CI & $P(\Delta>0)$ \\
\midrule
ZP              & 7.25 ± 1.41       & 0.26 & [0.11, 0.41] &  1.00       \\
PP              & 7.23 ± 1.47       & 0.27 & [0.12, 0.43] &  1.00    \\
SR              & 7.34 ± 1.37       & 0.18 & [0.03, 0.34] &  0.98       \\
\midrule
IPIR                & 7.53 ± 1.31       & -0.04 & [-0.19, 0.12] &  0.30      \\
EPIR                & 7.49 ± 1.34       & 0.0 & [-0.16, 0.15] &  0.49     \\
IPER                & 7.36 ± 1.37       & 0.11 & [-0.04, 0.23] &  0.92       \\
\midrule
\textbf{EPER} & \textbf{7.49 ± 1.35} & --  & -- & --         \\
\bottomrule
\end{tabular}
\caption{
Automatic evaluation results on the PerMPST dataset.
“Score” denotes the mean $\pm$ standard deviation of 10-point Likert ratings assigned by the LLM evaluator.
The table also reports results from a Bayesian ordinal regression model that treats Likert ratings as ordinal data. We report the posterior mean, 95\% credible interval, and posterior probability of the latent score difference $\Delta = \alpha_{\mathrm{EPER}} - \alpha_{method}$}
\label{tab:mpst-results}
\end{table}

Table~\ref{tab:comparison_winrates} presents the pairwise comparison results on the PerDOC dataset. For PerDOC, although comparisons with inconsistent judgments are discarded to mitigate position bias, the reported win rates are still based on several hundred consistent votes per comparison on average.\footnote{Across all 955 PerDOC pairs, an average of 499$\pm$101 comparisons were discarded due to disagreement, leaving approximately 455 consistent judgments per comparison (at least about 300 and up to about 670 valid votes).}

Our full configuration, EPER, outperforms all baseline models in win rates.
Among the baselines, PP is a strong in-context personalization method that directly incorporates past user preference history $H_u$ into the prompt.

The ZP baseline, which does not use any user preference information, achieves only a 20\% win rate against PP.
However, when using the same ZP outputs as initial plot for EPER refinement, the win rate increases to 67\%, representing a 47-point improvement. This demonstrates that even unpersonalized story drafts can be effectively personalized through our PREFINE architecture, highlighting its strength as a post-hoc personalization framework.

We also compare against Self-Refine (SR)~\cite{self-refine}, which uses a static rubric designed to improve general story quality. EPER achieves an 83\% win rate over SR, indicating that personalization-oriented refinement leads to significantly better alignment with individual user preferences.  

Finally, comparisons with model variants (IPIR, EPIR, IPER) highlight the importance of both expert-guided persona descriptions and user-specific rubric generation, with user-specific rubrics contributing the largest gains.

\subsection{Personalization Results on PerMPST}
\label{sec:mpst}
Table~\ref{tab:mpst-results} compares the proposed method with baselines and model variants.
To account for the ordinal nature of the 10-point Likert ratings, we analyze the results using a Bayesian model that estimates the posterior distribution of the latent score differences between methods, denoted as $\Delta$.

The results show that PREFINE (EPER) consistently achieves positive latent score differences ($\Delta > 0$) over all baseline methods, confirming improved personalization performance.

Although the analysis demonstrates the effectiveness of PREFINE, the absolute magnitude of the observed score differences appears modest.
This can be attributed to two factors:
(i) the stories in this dataset are relatively short (Appendix~\ref{appendix:token_dist}), leaving limited room for refinement; and (ii) LLM-based score evaluation tends to assign similar ratings across different systems, which can obscure performance differences~\cite{sahoo2025quantitativellmjudges}.

\begin{table}[t]
\centering
\footnotesize
\setlength{\tabcolsep}{4pt}
\begin{tabular}{lcccc}
\toprule
Method & Score & $\Delta$ & 95\% CI & $P(\Delta>0)$ \\
\midrule
PP    & 5.39 ± 1.98     & 2.54 & [1.83, 3.22] & 1.00  \\
SR       & 6.70 ± 1.65     & 1.21 & [0.55, 1.87] & 1.00  \\
\midrule
\textbf{EPER}& \textbf{7.82 ± 1.40} & -- & -- & -- \\
\bottomrule
\end{tabular}
\caption{
Human evaluation results on the PerMPST dataset.
“Score” indicates the mean ± standard deviation of 10-point Likert scores collected from 14 annotators across 4 story sets each ($n = 56$). The table also reports latent score differences $\Delta$ estimated using a Bayesian ordinal regression model, along with 95\% credible intervals and posterior probabilities relative to EPER.}

\label{tab:humanresult}
\end{table}

\subsection{Generalization to Another LLM}
\label{sec:generalization}
To examine whether PREFINE depends on the backbone model, we conducted additional experiments using \texttt{Mistral-7B}.
Using the same prompts and evaluation protocols as in the main experiments, PREFINE consistently outperforms the baselines on both PerDOC and PerMPST under automatic evaluation. The observed improvements are comparable to, and in some cases slightly stronger than, those obtained with \texttt{LLaMA-3-70B}, indicating that PREFINE generalizes beyond a specific LLM architecture. Detailed results are reported in Appendix~\ref{appendix:mistral-results}.

\subsection{Generated Story Quality}
In this section, we examine whether the personalization achieved by PREFINE affects the overall quality of the stories for general readers.

To manage evaluation cost, we created a subset of 200 story sets randomly sampled from each of the outputs generated on the PerDOC and PerMPST datasets. Each story was based on the same premise and was evaluated by \texttt{GPT-4o} using the six criteria defined in the \cite{hanna}. Scores were assigned on a scale from 1 to 10.

Notably, when averaged across all criteria, EPER achieves higher story quality than the Prompt-Persona (PP) baseline ($\Delta = 0.82$, 95\% CI [0.28, 1.34]) and comparable quality to Self-Refine (SR) ($\Delta = -0.30$, 95\% CI [-0.93, 0.32]).

These results show that EPER maintains overall story quality comparable to SR while achieving effective user-specific personalization. This suggests that the personalization gains are not merely a byproduct of improved general story quality, but that PREFINE improves stories along two distinct axes: general quality and user preference. A breakdown by evaluation aspect is provided in Appendix~\ref{appendix:general-rubric-quality}.

\begin{figure}[t]
  \centering
  \includegraphics[width=\columnwidth]{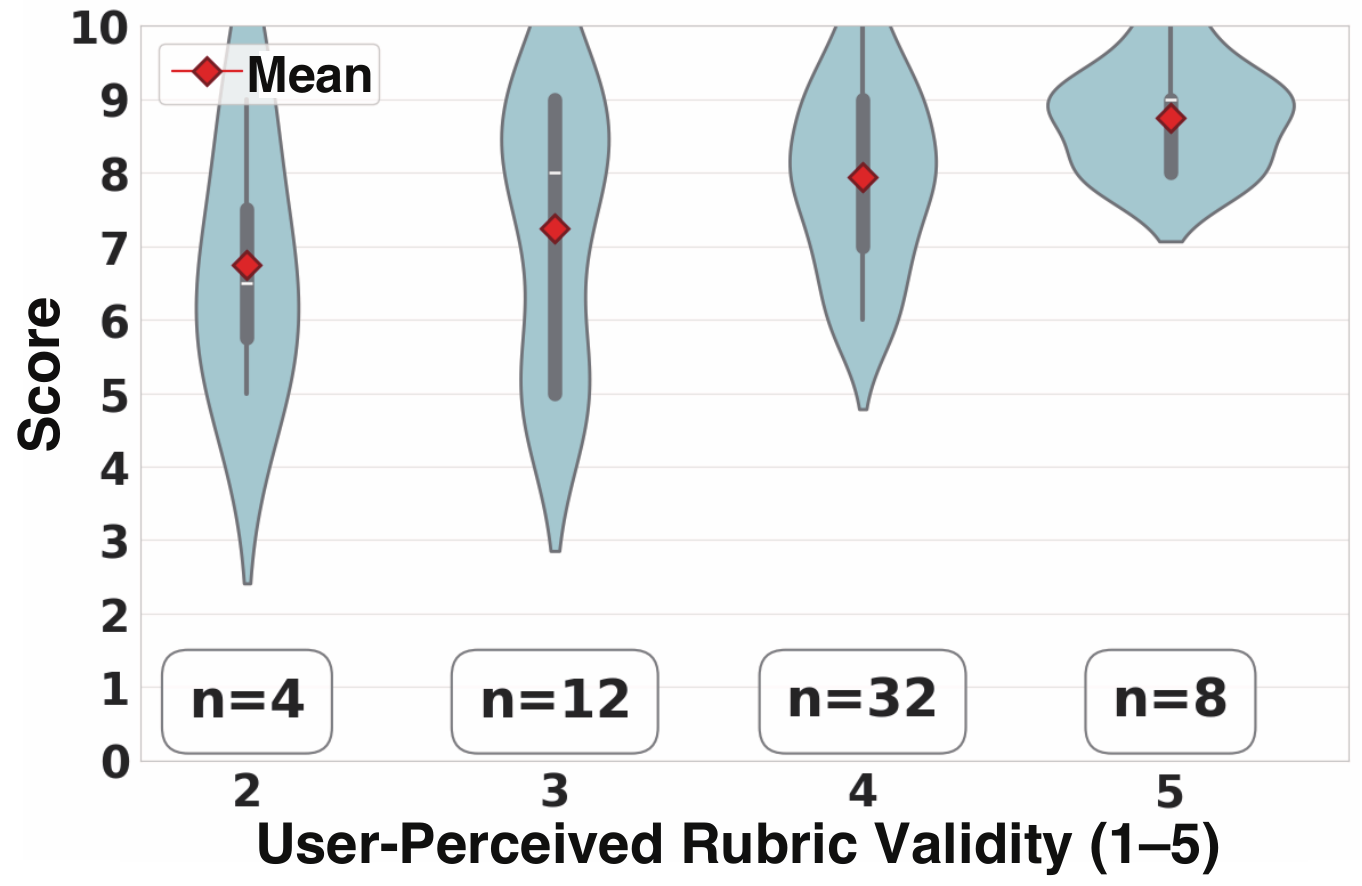}
  \caption{Relationship between users’ rubric suitability ratings and their preference scores for PREFINE-generated stories.}
  \label{fig:humaneval}
\end{figure}

\subsection{Human Evaluation and Analysis}
\paragraph{Human Preference Ratings}
We conducted a user study to evaluate PREFINE with actual users. The evaluation followed a similar design protocol (Section~\ref{sec:evaluation}) as used in PerMPST. 

This design was primarily motivated by the fact that the generated stories in PerMPST are shorter than those in PerDOC, which helps reduce cognitive load and allows annotators to maintain consistent evaluation quality.

Table~\ref{tab:humanresult} presents the results of the human evaluation. The findings demonstrate that our approach generates stories that better align with users’ actual preferences compared to both PP and SR. PREFINE achieved the highest average preference scores and the best average rank (lower is better; Avg. Rank — PREFINE 1.35, SR 1.98, PP 2.67). PREFINE consistently outperformed both PP and SR,
with posterior credible intervals for the latent score difference $\Delta$ lying entirely above zero, indicating a strong preference for our method.\footnote{The model incorporates random effects for participants and story premises to account for rater variability and premise-level differences.} Furthermore, PREFINE was effective across all premise sets used in our experiments. Detailed results, as well as supplementary analyses on inter-annotator agreement, are reported in Appendix \ref{appendix:per-premise}, \ref{appendix:IAA}.

These results are also consistent with the trends observed in automatic evaluation, reinforcing the effectiveness of our approach. Our findings highlight the value of user-specific rubrics for personalization. As future work, we plan to explore lighter human-in-the-loop methods that let users choose or adjust rubrics instead of engaging in the full critique–refine loop.

\begin{figure}[t]
  \centering
  \includegraphics[width=\columnwidth]{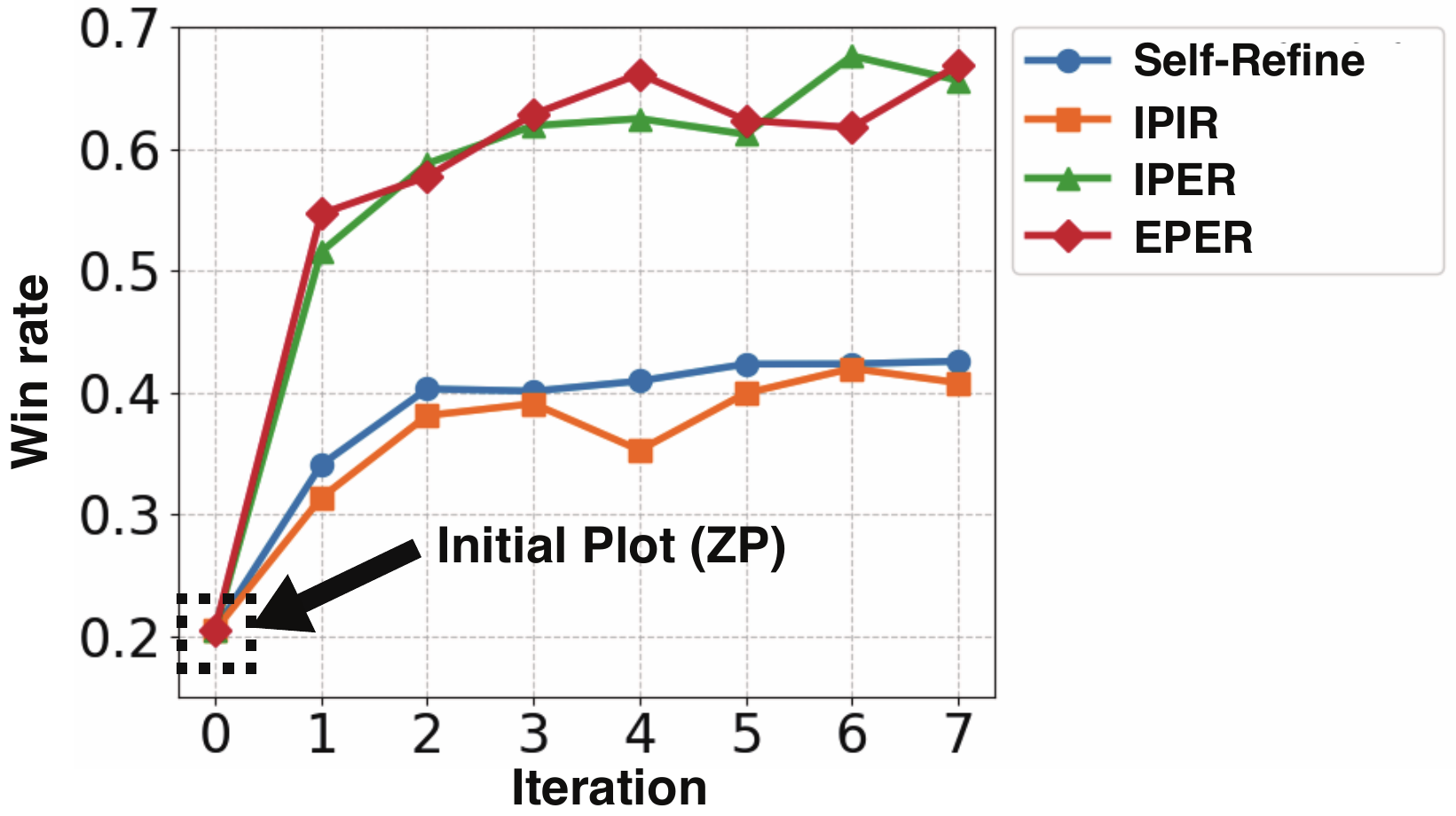}
  \caption{Win rate progression against Prompt-Persona (PP) across refinement iterations.}
  \label{fig:loop}
\end{figure}

\paragraph{User-Specific Rubric Quality}
\label{User-Specific Rubric Quality}
PREFINE incorporates a key component that generates and utilizes user-specific rubrics. In this section, we evaluate the quality of these rubrics based on real user feedback collected using a 5-point Likert scale. We also examine how perceived rubric quality relates to the effectiveness of personalization. 
As shown in Fig. \ref{fig:humaneval}, participants who perceived the rubrics as well aligned with their preferences tended to give higher preference scores\footnote{Note that this result may contain a slight evaluation bias, as some participants who rated the rubric highly may have consistently given higher scores to all methods. We briefly discuss the potential impact of such bias in Appendix~\ref{appendix:rubric_bias_analysis}.} to stories generated by PREFINE.

\section{Analysis}
\subsection{Feedback Loop for Personalization}

Figure~\ref{fig:loop} illustrates the change in win rates against PP across refinement iterations, comparing SR with PREFINE variants. Methods without user-specific rubrics, such as SR and IPIR, show limited improvement, whereas those with rubrics (EPER, IPER) exhibit consistent gains through iterations. These results indicate that user-specific rubrics play a key role in achieving effective personalization within the Critique-and-Refine framework.
\begin{table}[t]
\small
\centering
\begin{tabularx}{\columnwidth}{l|XXX} 
\toprule
\diagbox{A}{B} & PP & PEP & EPER \\
\midrule
\midrule
PP   & -- & \cellcolor{myblue!73} 0.27 & \cellcolor{myblue!96} 0.04 \\
PEP  & \cellcolor{myred!73} 0.73 & -- & \cellcolor{myblue!96} 0.04 \\
\textbf{EPER (Init: PEP)} & \cellcolor{myred!96} 0.96 & \cellcolor{myred!96} 0.96 & -- \\
\bottomrule
\end{tabularx}
\caption{Pairwise win rates between PP, PEP, and EPER (starting from PEP outputs) on the PerDOC dataset. EPER achieves further personalization even when initialized with already personalized PEP stories. See Appendix~\ref{appendix:aspect-winrate-perdoc-pep} for per-perspective results.}
\label{tab:my_comparison} 
\end{table}

\subsection{Effect on Personalized Outputs}
This section examines whether PREFINE can further improve outputs that are already personalized. Based on the PerDOC results (Sec.~\ref{sec:perdoc}), we introduce Prompt-Expert-Persona (PEP), which uses persona descriptions extracted by expert agents and outperforms Prompt-Persona (PP).

On PerDOC, PEP achieves a 73\% win rate over PP (Table~\ref{tab:my_comparison}), indicating stronger personalization.
Next, we use the output generated by PEP as the initial draft and apply PREFINE.
The results confirm that PREFINE can further enhance personalization, even when starting from highly personalized drafts.

We also conducted the same experiment on the PerMPST dataset, but no additional performance gains were observed. This outcome is consistent with the limitations discussed in Sec.~\ref{sec:mpst}, including short text lengths and evaluation bias. See Appendix~\ref{appendix:mpst-pep} for details.

\subsection{Length Bias in Automatic Evaluation}
Prior work reports that large language models (LLMs) tend to favor longer, more detailed outputs \cite{gu2025surveyllmasajudge, chatbotarena}.
We examine whether our LLM-based evaluator exhibits such length-dependent bias.

\begin{figure}[t]
  \centering
  \includegraphics[width=0.9\columnwidth]{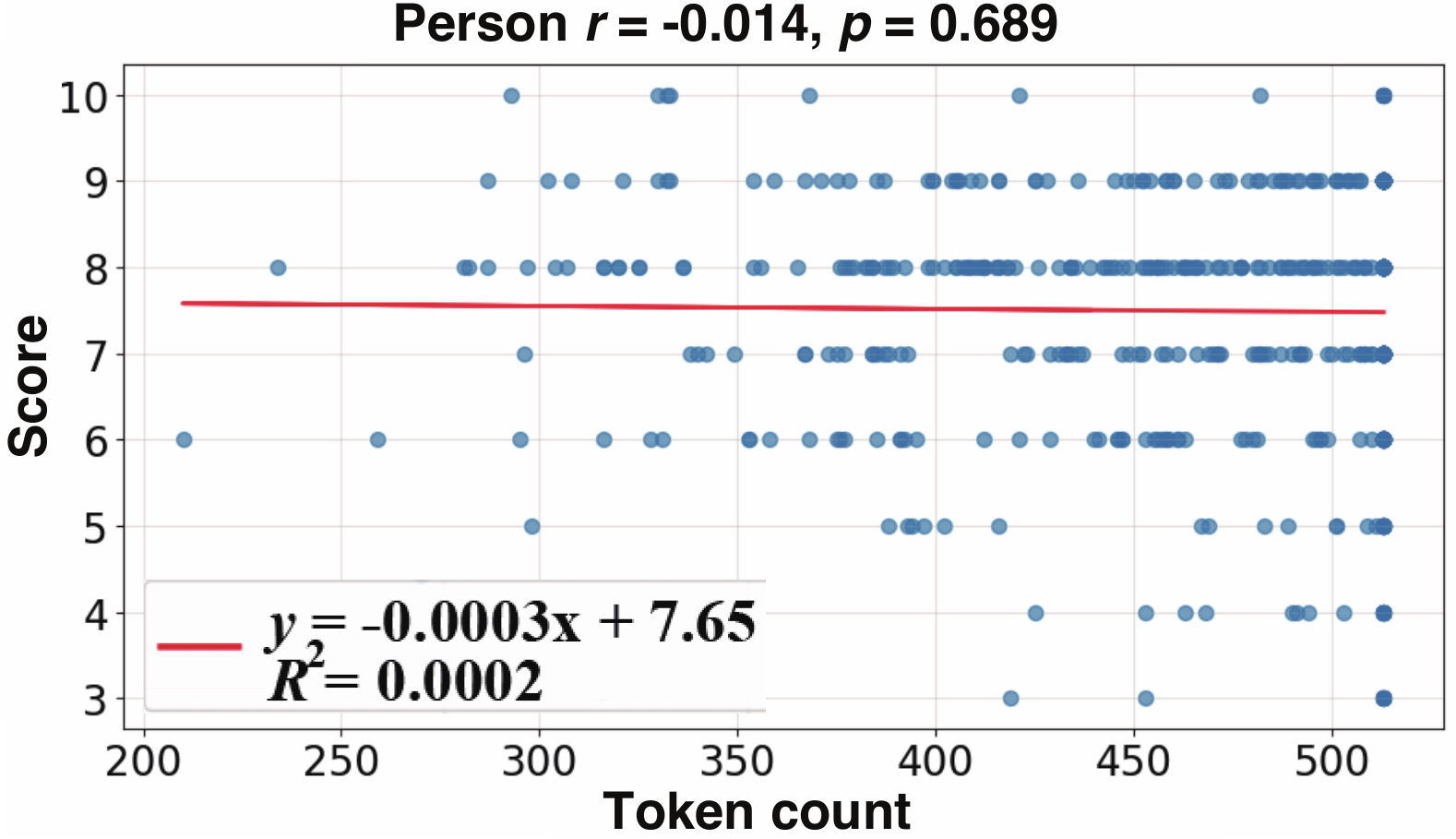}
  \caption{Correlation between token length and evaluation scores for EPER outputs.}
  \label{fig:token-corr}
\end{figure}

As shown in Fig.~\ref{fig:token-corr}, there is no significant correlation between token length and evaluation score for EPER outputs on the PerMPST dataset (Pearson $r=-0.014$, $p=0.689$). Likewise, on the PerDOC dataset, the influence of token length on win rate was also limited (See Appendix \ref{appendix:rel-perdoc-length}).

\section{Conclusion}
In this study, we proposed a novel personalized story generation method called PREFINE, which enables the creation of user-tailored stories without requiring explicit user feedback or additional training. PREFINE leverages a pseudo-user agent that mimics user preferences to generate user-specific rubrics. Based on the rubrics, the agent critiques and refines the story in multiple steps, gradually aligning with the user’s preferences.

Both automatic and human evaluations demonstrated that PREFINE achieves more effective personalization compared to the baselines. Furthermore, it was confirmed that PREFINE not only achieves successful personalization but also improves the general story quality to a level comparable to that of Self-Refine. The human evaluation results also suggest that the ability to construct rubrics that closely reflect user preferences plays a key role in the success of personalization within this framework.

The insights gained from PREFINE are not limited to story generation; we believe they can also serve as a foundation for personalized generation across other domains.

\section*{Limitations}
In this study, we evaluated PREFINE using two backbone models, \texttt{LLaMA-3-70B} and \texttt{Mistral-7B}, with a shared backbone across all agent roles within each experiment.
These results indicate that PREFINE is not tied to a single LLM architecture.
Exploring a broader design space, such as systematically varying model sizes or mixing different models across components, is beyond the scope of this work and left for future investigation.

While PREFINE enables training-free personalization, it incurs additional inference-time cost due to its iterative critique-and-refine process.
For example, on the PerDOC dataset, a single refinement loop increases the prompt by approximately 3{,}000 input tokens and produces about 1{,}000 additional output tokens per instance.
In principle, inference cost and latency could be further reduced by early stopping the feedback loop based on the feedback agent’s scores. However, in this work we intentionally disabled early stopping in order to analyze how the number of feedback steps affects personalization accuracy.
This design choice allows us to study the relationship between refinement depth and performance in a controlled manner.
Exploring adaptive stopping criteria is an important direction for future work.

This inference-time overhead reflects a trade-off between computation at inference time and the elimination of training-time costs, data collection requirements, and potential privacy concerns associated with fine-tuning.
In many realistic deployment scenarios, these training-time constraints are more restrictive, making PREFINE a practical choice despite its iterative nature.

Finally, the current implementation adopts a stateless design, reprocessing user histories and personas at each refinement step.
In practical systems, inference cost could be substantially reduced through stateful optimizations such as caching user personas and user-specific rubrics across interactions.

\bibliography{custom}

\appendix

\appendix
\section{Prompt Settings}
\label{appendix:promptsetting}
This section details the prompt templates used for each component in the PREFINE framework.

\subsection{Prompt for Initial Story Generation}
In the PerDOC setting, the initial story is generated following the same generation flow as in DOC \cite{perdoc}, which corresponds to the Zero-Persona (ZP) setting in this paper.

The setting for PerMPST is shown in Table~\ref{prompt:mpst-init-gen}.

\begin{table}[htbp]
\centering
\scriptsize
\begin{tabular}{p{0.95\linewidth}}
\toprule
\textbf{Prompt Template for Initial Story Generation on PerMPST ($\mathsf{prompt}_{\text{Init}}$)} \\
\midrule

You are a professional movie writer, skilled in crafting compelling and logically coherent synopsis.\\

Generate the continuation of the following movie synopsis.  \\
The synopsis should be between 10 and 13 sentences long and should not use bullet points.\\
Maintain a formal style consistent with official movie descriptions while ensuring logical coherence.\\
Preserve the given synopsis exactly as written. Begin your continuation immediately after the end of the premise, maintaining consistency in tone and content.\\
Do not modify, omit, or summarize the given synopsis.\\
Only output the completed synopsis without any additional commentary.\\

\texttt{\{premise\}} \\

\bottomrule
\end{tabular}
\caption{Prompt Template for Initial Story Generation on PerMPST ($\mathsf{prompt}_{\text{Init}}$)}
\label{prompt:mpst-init-gen}
\end{table}

\subsection{Prompt Template for Feedback from a Pseudo-User Agent}
\paragraph{For PerDOC}

We present the feedback prompt templates ($\mathsf{prompt}_{\text{feedback}}$) used for story critique and refinement in the PerDOC setting. These include the template for PREFINE’s full configuration, EPER (Table~\ref{prompt:doc-pseudo-agent-eper}), as well as those used in its variant models (IPIR(Table~\ref{prompt:doc-pseudo-agent-ipir}), EPIR(Table~\ref{prompt:doc-pseudo-agent-epir}), and IPER(Table~\ref{prompt:doc-pseudo-agent-iper})).

\begin{table}[htbp]
\centering
\scriptsize
\begin{tabular}{p{0.95\linewidth}}
\toprule
\textbf{Prompt Template Used by the Pseudo-User Agent in PREFINE (EPER): $\mathsf{prompt}_{\text{feedback}}$} \\
\midrule
You are a simulated literary critic who is thoroughly familiar with a specific user's narrative preferences. \\\\

[User Persona]\\
The following is a natural language summary of this user's storytelling preferences, derived from their past story evaluations. \\\\

\texttt{\{persona\_description\}}\\\\

Your task is to act from this user's perspective and provide feedback on a new story.\\
Your goal is to help improve the story so that it better satisfies the user's preferences under the aspect "\{aspect\}".\\\\

The following rubric has been generated to represent what this user considers important when evaluating stories in terms of "\{aspect\}".\\
Use this rubric to evaluate how well the story satisfies each criterion, and to suggest specific ways it can be improved.\\ 
Do not introduce new criteria or refer to the evaluation process.\\\\

[Rubric]  \\
\texttt{\{rubric\_list\}}\\\\

Each suggestion should aim to increase the score for that criterion.\\
Do not make any changes to the Premise. It is fixed and must remain unchanged in all your feedback.  \\
Do not summarize the plot.\\
In this scale, 5 represents a typical or average fulfillment of the criterion. \\Scores of 9 or 10 should be reserved for truly exceptional cases.\\
Keep your overall response under 200 tokens.\\\\

[Feedback Format]  \\
For each criterion:  \\
Criterion: \texttt{\{\{criterion\_text\}\}}\\
Score: X (1 = completely unsatisfactory, 10 = fully satisfies the criterion)\\
Explanation: ...\\
Suggestion: ...\\\\

[Story to Evaluate]  \\
\texttt{\{story\_plot\}}\\
\bottomrule
\end{tabular}
\caption{Prompt Template Used by the Pseudo-User Agent in PREFINE (EPER) on the PerDOC Dataset: $\mathsf{prompt}_{\text{feedback}}$}
\label{prompt:doc-pseudo-agent-eper}
\end{table}

\begin{table}[htbp]
\centering
\scriptsize
\begin{tabular}{p{0.95\linewidth}}
\toprule
\textbf{Prompt Template Used by the Pseudo-User Agent in PREFINE (IPIR): $\mathsf{prompt}_{\text{feedback}}$} \\
\midrule
You are a simulated literary critic who has internalized a specific user's narrative preferences based on their past story evaluations.\\\\

You are now asked to act from this user's perspective and provide feedback on a new story, reflecting what they would likely value or find lacking.\\
Your goal is to help improve the story so that it better aligns with what the user would likely prefer.\\\\

[Past Plot History]  \\
Below are two previously evaluated story plots (A and B), along with the user's selection and the evaluation aspect used at the time:\\\\

\texttt{\{user\_history\}}\\\\

[Selection Result]\\  
Aspect: \texttt{\{aspect\}}  \\
Choice: \texttt{\{choice\}}\\\\

The evaluation aspect is "{aspect}", and you are free to decide which elements matter most to this user within that aspect.  \\
Do not output a list of evaluation criteria or refer to the evaluation process.\\\\

Your response must include the following three parts, written in full sentences.  \\
Do not make any changes to the Premise. It is fixed and must remain unchanged in all your feedback.  \\
Do not summarize the plot, and do not explain the evaluation process.\\\\

Your feedback should be concise and focused, with no more than 8 sentences total (~200 tokens).\\\\

1. Positive Aspects  \\
2. Areas for Improvement  \\
3. Suggestions for Improvement \\\\

[Story to Evaluate]  \\
\texttt{\{story\_plot\}}\\
\bottomrule
\end{tabular}
\caption{Prompt Template Used by the Pseudo-User Agent in PREFINE (IPIR) on the PerDOC Dataset: $\mathsf{prompt}_{\text{feedback}}$}
\label{prompt:doc-pseudo-agent-ipir}
\end{table}

\begin{table}[htbp]
\centering
\scriptsize
\begin{tabular}{p{0.95\linewidth}}
\toprule
\textbf{Prompt Template Used by the Pseudo-User Agent in PREFINE (EPIR): $\mathsf{prompt}_{\text{feedback}}$} \\
\midrule
You are a simulated literary critic who is thoroughly familiar with a specific user's narrative preferences. \\\\

[User Persona]\\
The following is a natural language summary of this user's storytelling preferences, derived from their past story evaluations.\\\\

\texttt{\{persona\_description\}}\\\\

You are now asked to act from this user's perspective and provide feedback on a new story, reflecting what they would likely value or find lacking.\\
Your goal is to help improve the story so that it better aligns with what the user would likely prefer.\\\\

The evaluation aspect is "\texttt{\{aspect\}}", and you are free to decide which elements matter most to this user within that aspect.\\  
Do not output a list of evaluation criteria or refer to the evaluation process.\\\\

Your response must include the following three parts, written in full sentences.  \\
Do not make any changes to the Premise. It is fixed and must remain unchanged in all your feedback. \\ 
Do not summarize the plot, and do not explain the evaluation process.\\\\

Your feedback should be concise and focused, with no more than 8 sentences total (~200 tokens).\\\\

1. Positive Aspects  \\
2. Areas for Improvement \\ 
3. Suggestions for Improvement \\\\

[Story to Evaluate]  \\
\texttt{\{story\_plot\}}\\
\bottomrule
\end{tabular}
\caption{Prompt Template Used by the Pseudo-User Agent in PREFINE (EPIR) on the PerDOC Dataset: $\mathsf{prompt}_{\text{feedback}}$}
\label{prompt:doc-pseudo-agent-epir}
\end{table}

\begin{table}[htbp]
\centering
\scriptsize
\begin{tabular}{p{0.95\linewidth}}
\toprule
\textbf{Prompt Template Used by the Pseudo-User Agent in PREFINE (IPER): $\mathsf{prompt}_{\text{feedback}}$} \\
\midrule
You are a simulated literary critic who has internalized a specific user's narrative preferences based on their past story evaluations.\\\\

You are now asked to act from this user's perspective and provide feedback on a new story.\\
Your goal is to help improve the story so that it better satisfies the user's preferences under the aspect "{aspect}".\\\\

[Past Plot History]  \\
Below are two previously evaluated story plots (A and B), along with the user's selection and the evaluation aspect used at the time:\\\\

\texttt{\{user\_history\}}\\\\

[Selection Result]  \\
Aspect: \texttt{\{aspect\}}  \\
Choice: \texttt{\{choice\}}\\\\

The following rubric has been generated to represent what this user considers important when evaluating stories in terms of "{aspect}".  \\
Use this rubric to evaluate how well the story satisfies each criterion, and to suggest specific ways it can be improved.  \\
Do not introduce new criteria or refer to the evaluation process.\\\\

[Rubric]  \\
\texttt{\{rubric\_list\}} \\\\

Each suggestion should aim to increase the score for that criterion.\\
Do not make any changes to the Premise. It is fixed and must remain unchanged in all your feedback.  \\
Do not summarize the plot.\\
In this scale, 5 represents a typical or average fulfillment of the criterion. Scores of 9 or 10 should be reserved for truly exceptional cases.\\
Keep your overall response under 200 tokens.\\\\

[Feedback Format]  \\
For each criterion:  \\
Criterion: \texttt{\{\{criterion\_text\}\}} \\
Score: X (1 = completely unsatisfactory, 10 = fully satisfies the criterion)\\
Explanation: ...\\
Suggestion: ...\\\\

[Story to Evaluate]\\  
\texttt{\{story\_plot\}}\\
\bottomrule
\end{tabular}
\caption{Prompt Template Used by the Pseudo-User Agent in PREFINE (IPER) on the PerDOC Dataset: $\mathsf{prompt}_{\text{feedback}}$}
\label{prompt:doc-pseudo-agent-iper}
\end{table}

\paragraph{For PerMPST}
We present the feedback prompt templates ($\mathsf{prompt}_{\text{feedback}}$) used for story critique and refinement in the PerMPST setting. These include the template for PREFINE’s full configuration, EPER (Table~\ref{prompt:mpst-pseudo-agent-eper}), as well as those used in its variant models (IPIR(Table~\ref{prompt:mpst-pseudo-agent-ipir}), EPIR(Table~\ref{prompt:mpst-pseudo-agent-epir}), and IPER(Table~\ref{prompt:mpst-pseudo-agent-iper})).

\begin{table}[htbp]
\centering
\scriptsize
\begin{tabular}{p{0.95\linewidth}}
\toprule
\textbf{Prompt Template Used by the Pseudo-User Agent in PREFINE (EPER): $\mathsf{prompt}_{\text{feedback}}$} \\
\midrule
You are a simulated literary critic who has internalized a specific user's narrative preferences based on their past movie synopsis evaluations.\\\\

[User Persona]  \\
The following is a natural language summary of this user's storytelling preferences, derived from their prior evaluations of multiple movie synopses, each with an associated review and score.\\\\

\texttt{\{persona\_description\}}\\\\

You are now asked to act from this user's perspective and provide feedback on a new movie synopsis, reflecting what they would likely value or find lacking.  \\
Your goal is to help improve the synopsis so that it better aligns with what the user would likely prefer.\\\\

The following rubric has been generated to represent what this user considers important when evaluating movie synopses.  \\
Use this rubric to evaluate how well the given synopsis satisfies each criterion, and to suggest specific ways it can be improved.  \\
Do not introduce new criteria or refer to the evaluation process.\\\\

[Rubric]  \\
\texttt{\{rubric\_list\}} \\\\

Each suggestion should aim to increase the score for that criterion.  \\
Do not make any changes to the given premise. It is fixed and must remain unchanged in all your feedback.  \\
Premise: \texttt{\{premise\}}\\\\

Do not summarize the movie synopsis.  \\
In this scale, 5 represents a typical or average fulfillment of the criterion. Scores of 9 or 10 should be reserved for truly exceptional cases.  \\
Keep your overall response under 200 tokens.\\\\

[Feedback Format]  \\
For each criterion:  \\
Criterion: \texttt{\{\{criterion\_text\}\}}  \\
Score: X (1 = completely unsatisfactory, 10 = fully satisfies the criterion)  \\
Explanation: ...  \\
Suggestion: ... \\\\

[Movie Synopsis to Evaluate]  \\
\texttt{\{movie\_synopsis\}} \\
\bottomrule
\end{tabular}
\caption{Prompt Template Used by the Pseudo-User Agent in PREFINE (EPER) on the PerMPST Dataset: $\mathsf{prompt}_{\text{feedback}}$}
\label{prompt:mpst-pseudo-agent-eper}
\end{table}

\begin{table}[htbp]
\centering
\scriptsize
\begin{tabular}{p{0.95\linewidth}}
\toprule
\textbf{Prompt Template Used by the Pseudo-User Agent in PREFINE (IPIR): $\mathsf{prompt}_{\text{feedback}}$} \\
\midrule
You are a simulated literary critic who has internalized a specific user's narrative preferences based on their past movie synopsis evaluations.\\\\

You are now asked to act from this user's perspective and provide feedback on a new movie synopsis, reflecting what they would likely value or find lacking.\\
Your goal is to help improve the synopsis so that it better aligns with what the user would likely prefer.\\\\

[Past Synopsis History]\\
Below is a list of movie synopses you have previously reviewed, each with your review comments and a score from 1 (lowest) to 10 (highest).\\\\

\texttt{\{user\_history\}}\\\\

Use this information to infer what this user values in storytelling.\\
You may decide which elements to focus on based on your interpretation of their past evaluations.\\
Do not output a list of evaluation criteria or refer to the evaluation process.\\\\

Your response must include the following three parts, written in full sentences.\\
Do not make any changes to the given premise. It is fixed and must remain unchanged in all your feedback.\\
premise: \{premise\}\\\\

Do not summarize the synopsis, and do not explain the evaluation process.\\\\

Your feedback should be concise and focused, with no more than 8 sentences total (~200 tokens).\\\\

1. Positive Aspects\\
2. Areas for Improvement\\
3. Suggestions for Improvement\\\\

[Movie Synopsis to Evaluate]\\
\texttt{\{movie\_synopsis\}}\\
\bottomrule
\end{tabular}
\caption{Prompt Template Used by the Pseudo-User Agent in PREFINE (IPIR) on the PerMPST Dataset: $\mathsf{prompt}_{\text{feedback}}$}
\label{prompt:mpst-pseudo-agent-ipir}
\end{table}

\begin{table}[htbp]
\centering
\scriptsize
\begin{tabular}{p{0.95\linewidth}}
\toprule
\textbf{Prompt Template Used by the Pseudo-User Agent in PREFINE (EPIR): $\mathsf{prompt}_{\text{feedback}}$} \\
\midrule
You are a simulated literary critic who has internalized a specific user's narrative preferences based on their past movie synopsis evaluations.\\\\

[User Persona]\\
The following is a natural language summary of this user's storytelling preferences, derived from their prior evaluations of multiple movie synopses, each with an associated review and score.\\\\

\texttt{\{persona\_description\}}\\\\

You are now asked to act from this user's perspective and provide feedback on a new movie synopsis, reflecting what they would likely value or find lacking.\\
Your goal is to help improve the synopsis so that it better aligns with what the user would likely prefer.\\\\

Do not output a list of evaluation criteria or refer to the evaluation process.\\\\

Your response must include the following three parts, written in full sentences.\\
Do not make any changes to the given premise. It is fixed and must remain unchanged in all your feedback.\\
Premise: \texttt{\{premise\}}\\\\

Do not summarize the synopsis, and do not explain the evaluation process.\\\\

Your feedback should be concise and focused, with no more than 8 sentences total (~200 tokens).\\\\

1. Positive Aspects  \\
2. Areas for Improvement \\ 
3. Suggestions for Improvement \\\\

[Movie Synopsis to Evaluate]  \\
\texttt{\{movie\_synopsis\}}\\
\bottomrule
\end{tabular}
\caption{Prompt Template Used by the Pseudo-User Agent in PREFINE (EPIR) on the PerMPST Dataset: $\mathsf{prompt}_{\text{feedback}}$}
\label{prompt:mpst-pseudo-agent-epir}
\end{table}

\begin{table}[htbp]
\centering
\scriptsize
\begin{tabular}{p{0.95\linewidth}}
\toprule
\textbf{Prompt Template Used by the Pseudo-User Agent in PREFINE (IPER): $\mathsf{prompt}_{\text{feedback}}$} \\
\midrule
You are a simulated literary critic who has internalized a specific user's narrative preferences based on their past movie synopsis evaluations.\\\\

You are now asked to act from this user's perspective and provide feedback on a new movie synopsis, reflecting what they would likely value or find lacking.\\
Your goal is to help improve the synopsis so that it better aligns with what the user would likely prefer.\\\\

[Past Synopsis History]  \\
Below is a list of movie synopses you have previously reviewed, each with your review comments and a score from 1 (lowest) to 10 (highest).\\\\

\texttt{\{user\_history\}}\\\\

The following rubric has been generated to represent what this user considers important when evaluating movie synopses.  \\
Use this rubric to evaluate how well the given synopsis satisfies each criterion, and to suggest specific ways it can be improved. \\ 
Do not introduce new criteria or refer to the evaluation process.\\\\

[Rubric]  \\
\texttt{\{rubric\_list\}}\\\\

Each suggestion should aim to increase the score for that criterion.  \\
Do not make any changes to the Premise. It is fixed and must remain unchanged in all your feedback.\\
premise: \texttt{\{premise\}}\\\\

Do not summarize the movie synopsis.  \\
In this scale, 5 represents a typical or average fulfillment of the criterion. Scores of 9 or 10 should be reserved for truly exceptional cases.  \\
Keep your overall response under 200 tokens.\\\\

[Feedback Format]  \\
For each criterion:  \\
Criterion: \texttt{\{\{criterion\_text\}\}}  \\
Score: X (1 = completely unsatisfactory, 10 = fully satisfies the criterion)  \\
Explanation: ...  \\
Suggestion: ... \\\\

[Movie Synopsis to Evaluate]  \\
\texttt{\{movie\_synopsis\}}\\
\bottomrule
\end{tabular}
\caption{Prompt Template Used by the Pseudo-User Agent in PREFINE (IPER) on the PerMPST Dataset: $\mathsf{prompt}_{\text{feedback}}$}
\label{prompt:mpst-pseudo-agent-iper}
\end{table}

\subsection{Generating Explicit Personas ($EP$) via Expert LLM Agents}
\label{appnedix:ExpertLLMAgents}
Here, we present the prompt used to extract a user's explicit persona ($EP$) using an expert agent constructed like \cite{park2024generativeagentsimulations1000}, serving as a simulated pseudo-user agent.
The prompt used for the PerDOC setting is shown in Table~\ref{prompt:doc-ep-extraction}, while the one for PerMPST is shown in Table~\ref{prompt:mpst-ep-extraction}.

\begin{table}[htbp]
\centering
\scriptsize
\begin{tabular}{p{0.95\linewidth}}
\toprule
\textbf{Prompt template for deriving explicit user personas ($EP$) using an expert LLM agent on the PerDOC dataset.} \\
\midrule
Imagine you are an expert psychologist (with a PhD) taking notes while analyzing an individual's story preferences. You have been given two story plots, a question based on a specific evaluation criterion (Aspect) regarding them, and an individual's binary choice as a response to the question. Write observations\/reflections about the individual's personality traits, cognitive style, emotional tendencies, and psychological motivations based on their preference. (You should make more than 5 observations and fewer than 10. Choose the number that makes sense given the depth of the story plots and the individual's choice.)\\\\
\texttt{\{user\_preference\}} \\
\text{[Aspect]} \\
\texttt{\{aspect\}} \\
\text{[Preference]} \\
\texttt{\{user\_preference\_answer\}} \\\\
Follow the instructions and write only observations and reflections. Do not include anything else. Do not use 'plot A,' 'plot B,' or the word 'plot'.\\
\bottomrule
\end{tabular}
\caption{Prompt template for deriving explicit user personas ($EP$) using an expert LLM agent on the PerDOC dataset.}
\label{prompt:doc-ep-extraction}
\end{table}

\begin{table}[htbp]
\centering
\scriptsize
\begin{tabular}{p{0.95\linewidth}}
\toprule
\textbf{Prompt template for deriving explicit user personas ($EP$) using an expert LLM agent on the PerMPST dataset.} \\
\midrule
Imagine you are an expert psychologist (with a PhD) taking notes while analyzing an individual's story preferences. You have been given information about a reviewer's preferences, including multiple movie plots, as well as their review and score for each movie plot, ranging from 1 (lowest) to 10 (highest). Write observations\/reflections about the individual's personality traits, cognitive style, emotional tendencies, and psychological motivations based on their preferences. (You should make more than 5 observations and fewer than 10. Choose a number that makes sense given the depth of the story plots and the individual's choices.)\\\\

\texttt{\{user\_preference\}}\\\\

Follow the instructions and write only observations and reflections. Do not include anything else. Do not use 'plot 0,' 'plot 1,' 'plot 2,' 'plot 3,' or the word 'plot'.\\
\bottomrule
\end{tabular}
\caption{Prompt template for deriving explicit user personas ($EP$) using an expert LLM agent on the PerMPST dataset.}
\label{prompt:mpst-ep-extraction}
\end{table}

\subsection{Constructing User-Specific Rubrics from Interaction History}
We present the prompt templates used to construct the User-Specific Rubric for \textit{Explicit Rubric (ER)}.
Different prompts are used depending on whether the pseudo-user agent is based on the \textit{Explicit Persona (EP)} or the \textit{Implicit Persona (IP)}.
For the PerDOC dataset, the prompt for the EP-based agent is shown in Table~\ref{prompt:doc-rubric-extraction-ep}, and the one for the IP-based agent is shown in Table~\ref{prompt:doc-rubric-extraction-ip}.
Similarly, for the PerMPST dataset, the EP-based prompt is presented in Table~\ref{prompt:mpst-rubric-extraction-ep}, and the IP-based prompt in Table~\ref{prompt:mpst-rubric-extraction-ip}.

\begin{table}[htbp]
\centering
\scriptsize
\begin{tabular}{p{0.95\linewidth}}
\toprule
\textbf{Prompt template for generating user-specific rubrics from interaction history in the PerDOC (EP version).} \\
\midrule
You are a simulated literary critic aligned with a specific user's narrative preferences, derived from their past story evaluations.\\\\

[User Persona]\\
\texttt{\{user\_history\}}\\\\

Based on this persona, construct a personalized rubric for how the user likely interprets the narrative aspect "\texttt{\{aspect\}}".\\\\

List 3 to 5 specific criteria that this user would likely consider important when evaluating this aspect.\\
Each criterion should be phrased as a short, standalone natural language statement.\\
These criteria will later guide your feedback on the story.\\
Do not explain or elaborate—only list the criteria clearly.\\\\

[Your Rubric]\\
\bottomrule
\end{tabular}
\caption{Prompt template for generating user-specific rubrics from interaction history in the PerDOC (\textit{Explicit Persona (EP)} version).}
\label{prompt:doc-rubric-extraction-ep}
\end{table}

\begin{table}[htbp]
\centering
\scriptsize
\begin{tabular}{p{0.95\linewidth}}
\toprule
\textbf{Prompt template for generating user-specific rubrics from interaction history in the PerDOC (IP version).} \\
\midrule
You are a simulated literary critic who has internalized a specific user's narrative preferences through prior story evaluations.\\\\

[Past Plot History]\\
Two previously evaluated story plots (A and B) are shown, along with the user's chosen story and the evaluation aspect at the time:\\\\

\texttt{\{user\_history\}}\\\\

[Selection Result]\\
Aspect: \texttt{\{aspect\}}\\
Choice: \texttt{\{choice\}}\\\\

Based on this information, construct a personalized rubric for how the user likely interprets the narrative aspect "\texttt{\{aspect\}}".\\\\

List 3 to 5 specific criteria that this user would likely consider important when evaluating this aspect.\\
Each criterion should be phrased as a short, standalone natural language statement.\\
These criteria will later guide your feedback on the story.\\
Do not explain or elaborate—only list the criteria clearly.\\\\

[Your Rubric]\\
\bottomrule
\end{tabular}
\caption{Prompt template for generating user-specific rubrics from interaction history in the PerDOC (\textit{Implicit Persona (IP)} version).}
\label{prompt:doc-rubric-extraction-ip}
\end{table}

\begin{table}[htbp]
\centering
\scriptsize
\begin{tabular}{p{0.95\linewidth}}
\toprule
\textbf{Prompt template for generating user-specific rubrics from interaction history in the PerMPST (EP version).} \\
\midrule
You are a simulated literary critic aligned with a specific user's narrative preferences, derived from their past movie synopsis evaluations.\\\\

[User Persona]\\
\texttt{\{user\_history\}}\\\\

Based on this Persona, construct a personalized rubric that reflects the characteristics of synopses this user tends to prefer.\\\\

List 3 to 5 specific criteria that this user would likely consider important when evaluating synopses.\\
Each criterion should be phrased as a short, standalone natural language statement.\\
These criteria will later guide your feedback on the story.\\
Do not explain or elaborate—only list the criteria clearly.\\\\

[Your Rubric]\\
\bottomrule
\end{tabular}
\caption{Prompt template for generating user-specific rubrics from interaction history in the PerMPST (\textit{Explicit Persona (EP)} version).}
\label{prompt:mpst-rubric-extraction-ep}
\end{table}

\begin{table}[htbp]
\centering
\scriptsize
\begin{tabular}{p{0.95\linewidth}}
\toprule
\textbf{Prompt template for generating user-specific rubrics from interaction history in the PerMPST (IP version).} \\
\midrule
You are a simulated literary critic who has internalized a specific user's narrative preferences through prior movie synopsis evaluations.\\\\

[Past Plot History]\\
You are given a reviewer's preferences based on several previously rated movie plots.\\
Each preference includes a plot synopsis, a short review, and a numeric score from 1 (lowest) to 10 (highest), indicating how much the reviewer liked the movie synopsis.\\\\

\texttt{\{user\_history\}}\\\\

Based on this information, construct a personalized rubric that reflects the characteristics of synopses this user tends to prefer.\\\\

List 3 to 5 specific criteria that this user would likely consider important when evaluating synopses.\\
Each criterion should be phrased as a short, standalone natural language statement.\\
These criteria will later guide your feedback on the story.\\
Do not explain or elaborate—only list the criteria clearly.\\\\

[Your Rubric]\\
\bottomrule
\end{tabular}
\caption{Prompt template for generating user-specific rubrics from interaction history in the PerMPST (\textit{Implicit Persona (IP)} version).}
\label{prompt:mpst-rubric-extraction-ip}
\end{table}

\subsection{Prompt Template: Refinement Agent}

Here, we present the refinement prompt($\mathsf{prompt}_{\text{refine}}$) used by the refinement agent $\mathcal{M}_{\text{refine}}$ for personalization.
The prompt template for the PerDOC dataset is shown in Table\ref{prompt:refinement-agent-doc}, and the one for the PerMPST dataset is shown in Table\ref{prompt:refinement-agent-mpst}

\begin{table}[htbp]
\centering
\scriptsize
\begin{tabular}{p{0.95\linewidth}}
\toprule
\textbf{Refinement prompt template $\mathsf{prompt}_{\text{refine}}$ used by the refinement agent in the PerDOC setting.} \\
\midrule
You are a professional fiction editor. Your task is to refine a story plot based on the given feedback while strictly preserving the structural format.\\
You may rewrite, add, modify, or delete parts of the text as needed to improve the story based on the feedback.  \\\\

\text{[Feedback Start]}\\
\texttt{\{feedback\}}\\
\text{[Feedback End]}\\\\

[Structural Template Start]\\\\

Premise:\\
The fundamental premise of the story.\\
**Do not change the Premise under any circumstances.**\\\\

Setting:\\
Information about the story’s setting.\\
Include only the setting description that begins with 'The story is set in'.\\\\

Characters:\\
A list of main characters, including their names and portraits.\\\\

Outline:\\
A structured summary of the story.\\
You must write exactly **FOUR top-level items**, numbered 1 to 4.\\
Each item must contain at least one sub-point, and may include up to four (a-d).\\
Using fewer than four is acceptable if appropriate.\\\\

[Structural Template End]\\\\

Output Requirements:\\
1. **The refined plot must be between 500-550 tokens in length.**\\
2. The final text must be complete, coherent, and self-contained.\\
3. Return only the refined plot using the exact structure. Additional explanations or comments are not allowed.\\
4. **The Outline section must contain EXACTLY four top-level items (1 to 4), and each must include AT LEAST one sub-point (a–d).** You may include up to four sub-points per item if appropriate.\\
\bottomrule
\end{tabular}
\caption{Refinement prompt template $\mathsf{prompt}_{\text{refine}}$ used by the refinement agent in the PerDOC setting.}
\label{prompt:refinement-agent-doc}
\end{table}

\begin{table}[htbp]
\centering
\scriptsize
\begin{tabular}{p{0.95\linewidth}}
\toprule
\textbf{Refinement prompt template $\mathsf{prompt}_{\text{refine}}$ used by the refinement agent in the PerMPST setting.} \\
\midrule
This task requires refining a plot based on the provided feedback. \\
The refined plot must incorporate the provided feedback. \\\\

[Feedback Start]\\
\texttt{\{feedback\}}\\
\text{[Feedback End]}\\\\

---\\\\

[Instructions for Refinement]\\
- Accurately apply the feedback while maintaining the essence of the original plot.\\\\

[Output Requirements]\\
- Strictly follow the structural template.\\
- Modifications to the Premise are not permitted. premise -\> \texttt{\{premise\}}\\
- Apply the necessary modifications while ensuring consistency and logical coherence.\\
- Please keep the number of words similar to the original plot.\\
- Include only the refined plot. Additional explanations or comments are not required.\\
\bottomrule
\end{tabular}
\caption{Refinement prompt template $\mathsf{prompt}_{\text{refine}}$ used by the refinement agent in the PerMPST setting.}
\label{prompt:refinement-agent-mpst}
\end{table}

\section{Details of the Evaluation}

\subsection{LLM Evaluator Selection and Validation}
\label{appendix:evaluator_selection}
It has been reported that LLM-based judges, when provided with a user's interaction history $H_u$, can partially predict the user's preference judgments for stories (e.g., binary preference choices or rating-based evaluations) \cite{perse}. PerSE \cite{perse} is an LLM designed to align closely with user preferences, and is fine-tuned on llama-based models using either the PerDOC or PerMPST datasets.

Following the procedure in PerSE \cite{perse}, we fine-tuned a model based on LLaMA3\footnote{\texttt{meta-llama/Llama-3.1-8B-Instruct}} to construct PerSE-Llama3-8B and gemma3\footnote{\texttt{google/gemma-3-12b-it}} to construct PerSE-gemma3-12B

We then evaluated its alignment with human judgments using the evaluation set provided by \cite{perse}, comparing our model's performance against other existing LLMs, such as GPT-4o. While \cite{perse} used Llama2\footnote{\texttt{meta-llama/Llama-2-7b-chat-hf}, \texttt{meta-llama/Llama-2-13b-chat-hf}} as the backbone, we adopted Llama3 to improve overall performance.

The results are shown in Table~\ref{tab:llm_evalater_perdoc},\ref{tab:llm_evalater_permpst}. On the PerDOC dataset, our fine-tuned model (PerSE-Llama3-8B) achieved the highest accuracy (0.633), outperforming other models. On the other hand, in the PerMPST dataset, GPT-4o showed the highest correlation with human preferences, as measured by Pearson, Spearman, and Kendall correlation coefficients.

\paragraph{Judge-model sensitivity}
PerSE-Llama3-8B is a strong judge that aligns well with the human preference labels, but it shares a family with the generator, raising concerns about the same family stylistic bias \cite{llm-fam-bias}.

Accordingly, we compare the judgments made by PerSE–Gemma3-12B, which showed the second-highest agreement with human annotations, to examine potential concerns about style bias.

For validation, we use the pairwise comparison results between EPER and Prompt-Persona (PP), as PP serves as a strong competing baseline against EPER.

In the same query comparison (ties excluded), the between-judge win-rate difference was +5.59 percentage points, with a 90\% confidence interval (CI) of [2.28, 8.90] percentage points. Under a pre-specified equivalence margin of $\delta$ = 10 percentage points (pp), the two one-sided tests (TOST) procedure supported equivalence. As a result, the potential influence of stylistic bias arising from using models within the same family appears to be limited.

Accordingly, we adopt PerSE-Llama3-8B as the LLM judge for story evaluation in the PerDOC setting, and GPT-4o\footnote{\texttt{gpt-4o-2024-08-06}} as the LLM judge for the PerMPST setting.

\begin{table*}[htbp]
\centering
\scriptsize
\begin{tabular}{lccccc|c}
\toprule
 & \textbf{Interestingness} & \textbf{Adaptability} & \textbf{Surprise} & \textbf{Character} & \textbf{Ending} & \textbf{Average} \\
\midrule
GPT-4(from~\cite{perse}) & 0.502 & 0.496 & 0.596 & 0.506 & 0.543 & 0.529 \\
GPT-4o & 0.462 & 0.473 & 0.491 & 0.463 & 0.497 & 0.476 \\
PerSE-Llama2 (7B, from~\cite{perse}) & 0.572 & 0.565 & 0.619 & 0.565 & 0.560 & 0.576 \\
PerSE-Llama2 (13B, from~\cite{perse}) & \textbf{0.621} & 0.570 & 0.616 & 0.607 & 0.597 & 0.602 \\
PerSE-gemma3 (12B) & 0.590 & 0.625 & 0.553 & 0.586 & 0.590 &  0.589 \\
\textbf{PerSE-Llama3 (8B)} & 0.594 & \textbf{0.643} & \textbf{0.683} & \textbf{0.617} & \textbf{0.642} & \textbf{0.633} \\
\bottomrule
\end{tabular}
\caption{Evaluation accuracy across five aspects on the PerDOC dataset. 
Each model predicts user preferences based on the interaction history $H_u$, with the number of interactions fixed at $K=1$, consistent with the setting in \cite{perse}. 
Bold values indicate the highest score in each aspect. 
Results for GPT-4 and PerSE-Llama2 models are reproduced from \cite{perse}; all others are newly evaluated.}
\label{tab:llm_evalater_perdoc}
\end{table*}

\begin{table*}[htbp]
\centering
\begin{tabular}{lccc}
\toprule
\textbf{} & \textbf{Pearson} & \textbf{Spearman} & \textbf{Kendall} \\
\midrule
GPT-4 (from~\cite{perse}) & 0.315 & 0.312 & 0.253 \\
\textbf{GPT-4o} & \textbf{0.3831} & \textbf{0.4065} & \textbf{0.3239} \\ 
PerSE-Llama2 (7B, from~\cite{perse}) & 0.307 & 0.329 & 0.263 \\
PerSE-Llama2 (13B, from~\cite{perse}) & 0.345 & 0.368 &  0.293\\
PerSE-gemma3 (12B) & 0.2789 & 0.3066 & 0.2460 \\
PerSE-Llama3 (8B) & 0.3647 & 0.3817 & 0.3074 \\
\bottomrule
\end{tabular}
\caption{Correlation coefficients (Pearson, Spearman, Kendall) between predicted and human-annotated ground-truth scores in the PerMPST dataset. The results for GPT-4 and PerSE (based on LLaMA2) are taken directly from the original study \cite{perse}. Note that these values correspond to the condition where the number of interaction histories $K=3$ is provided as input. In contrast, the results for GPT-4o and our Llama3-based PerSE were computed under the $K=4$ condition. However, according to Figure 3 in \cite{perse}, the performance of Llama2-based PerSE remains largely unchanged between $K=3$ and $K=4$, and GPT-4o consistently shows the highest correlation with human judgments. Bold values indicate the highest correlation score. Based on these findings, we adopt GPT-4o as the LLM judge for the PerMPST setting in our experiments.
}
\label{tab:llm_evalater_permpst}
\end{table*}

\subsection{Human Evaluation Protocol}
\label{appendix:human_eval}
\begin{figure*}[htbp]
  \centering
  \includegraphics[width=\textwidth]{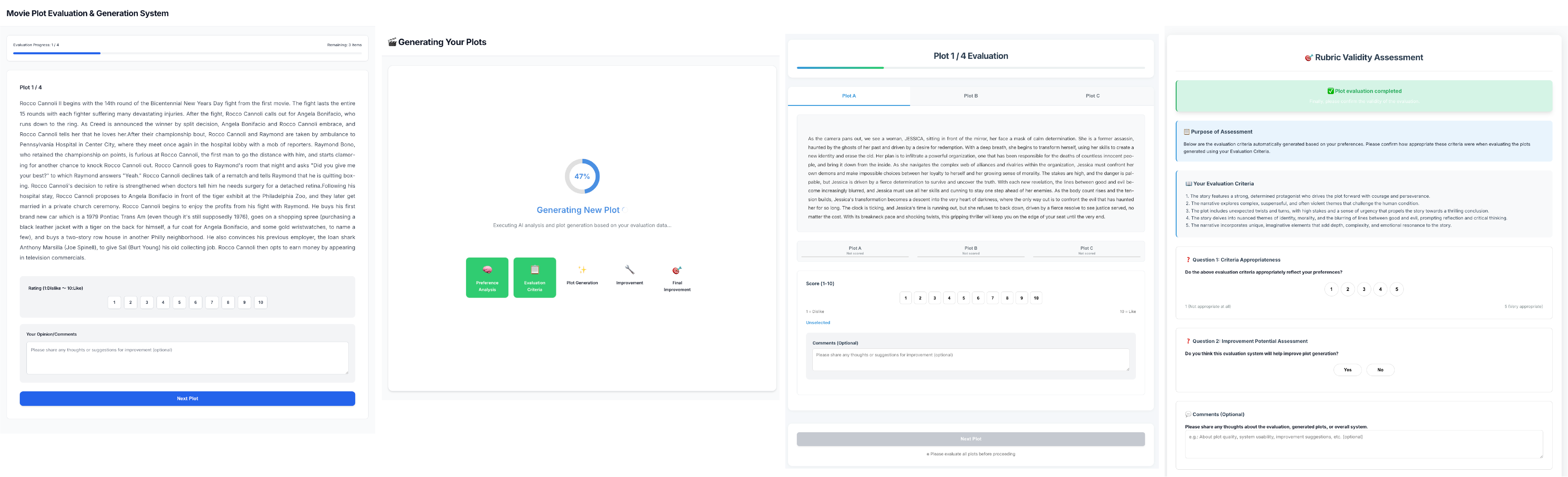}
  \caption{Interface of the web application used in the user study. The application enables end-to-end evaluation by collecting user preferences (via synopsis ratings and comments), presenting generated stories for evaluation, and collecting user feedback on personalized user-specific rubrics.}
  \label{fig:webapp}
\end{figure*}

We conducted a user study to evaluate the proposed method with real participants. The entire evaluation was performed end-to-end through a custom-built web application, which handled user preference collection, story generation, and user evaluation. A screenshot of the application interface is shown in Figure~\ref{fig:webapp}.

\paragraph{Preference Collection.}
At the beginning of the study, each participant was shown four movie synopses, manually selected from the PerMPST dataset to ensure genre diversity. These four synopses were identical across all participants. For each summary, participants were asked to rate their preference on a 10-point Likert scale (1 = dislike, 10 = like) and provide review comments. The resulting set of four $(synopsis, score, comment)$ tuples was used as the user's interaction history $H_u$.

\paragraph{Story Generation.}
For each participant, we used four predefined premises \(c_1,\dots,c_4\), which were identical across all participants.
Given the participant's interaction history \(H_u\), one story was generated for each premise using three different methods—Prompt\textendash Persona (PP), Self\textendash Refine (SR), and our proposed method EPER—resulting in \(3 \times 4 = 12\) stories per participant.
Before presentation, the three stories associated with each premise were randomly shuffled.

\paragraph{Story Evaluation.}
Participants rated each story on a 10-point scale (1 = dislike, 10 = like). In cases where multiple stories received the same score, participants were asked to assign a ranking (1 to 3) to indicate relative preference. This process was repeated for all four premises.

\paragraph{Rubric Evaluation.}
After the story evaluation, participants were shown the rubric generated by EPER for their preferences. They were then asked to evaluate how well the rubric reflected their preference criteria using a 5-point Likert scale (1 = not appropriate, 5 = very appropriate).

\paragraph{Participants.}
A total of 14 participants were recruited from graduate programs (master’s and doctoral levels) at our university. For non-English speakers (e.g., Japanese-speaking participants), the stories and rubrics were translated into their native language using GPT-4o, which has demonstrated strong translation performance.  
The quality and fidelity of these translations were verified by both the authors and fluent English speakers to ensure that the translated texts preserved the original intent and nuances.  
The entire procedure, including instructions and story generation time, was completed within approximately 60 minutes per participant. Participants received a monetary compensation of 1,500 JPY for their participation, which is consistent with standard compensation rates for similar user studies in Japan.

\subsection{Per-Premise Human Evaluation Results}
\label{appendix:per-premise}
In the human evaluation, we used four different story premises.
For each premise, multiple system outputs were presented, and we adopted a within-subject comparative design in which the same participants compared these outputs.

Table~\ref{tab:per-premise-human} reports the average preference scores for each premise.
Across all four premises, PREFINE (EPER) consistently outperforms both Prompt-Persona and Self-Refine,
confirming that the observed improvements are not driven by any specific premise.
These results indicate that the effectiveness of PREFINE does not depend on particular story settings,
but instead generalizes across different premises by producing outputs that are stably aligned with user preferences.

Overall, we believe that the evaluation scale using four distinct premises is sufficient to support the conclusion
that PREFINE better aligns with user preferences than the baseline methods.

\begin{table}[t]
\centering
\small
\setlength{\tabcolsep}{6pt}
\begin{tabular}{c l c c}
\toprule
Premise ID & Method & Mean & Std. \\
\midrule
\multirow{3}{*}{0}
 & PREFINE (EPER)      & 8.214 & 0.975 \\
 & Self-Refine         & 5.857 & 1.562 \\
 & Prompt-Persona     & 5.714 & 2.164 \\
\midrule
\multirow{3}{*}{1}
 & PREFINE (EPER)      & 7.500 & 1.506 \\
 & Self-Refine         & 6.929 & 1.328 \\
 & Prompt-Persona     & 4.786 & 1.762 \\
\midrule
\multirow{3}{*}{2}
 & PREFINE (EPER)      & 7.571 & 1.284 \\
 & Self-Refine         & 6.786 & 2.045 \\
 & Prompt-Persona     & 5.500 & 1.829 \\
\midrule
\multirow{3}{*}{3}
 & PREFINE (EPER)      & 8.000 & 1.754 \\
 & Self-Refine         & 7.214 & 1.424 \\
 & Prompt-Persona     & 5.571 & 2.209 \\
\bottomrule
\end{tabular}
\caption{
Per-premise human evaluation results on the PerMPST dataset.
Mean and standard deviation of 10-point Likert preference scores are reported for each method.
Across all premises, PREFINE (EPER) consistently achieves the highest average scores.
}
\label{tab:per-premise-human}
\end{table}

\subsection{Annotator Agreement}
\label{appendix:IAA}
In this study, stories are personalized for each annotator based on their individual interaction history and the given story premise.
As a result, multiple annotators do not evaluate the same generated output.
For this reason, standard inter-annotator agreement measures such as Fleiss’ $\kappa$,
which assume that multiple annotators assign labels to the same instances,
are not directly applicable in our evaluation setting.

On the other hand, each participant evaluates four sets of stories corresponding to their own premise,
and within each set assigns scores to outputs generated by multiple methods.
This design allows us to compute, for each annotator, the mean and variance of the scores assigned to each method.
By aggregating these rater-level statistics across annotators, we can analyze overall evaluation trends and variability across methods.

Table~\ref{tab:rater-level-stats} reports, for each method,
the mean of rater-level mean scores and the mean of rater-level standard deviations.
The results show that PREFINE (EPER) consistently receives higher ratings across annotators,
providing supplementary evidence for the stability and robustness of the human evaluation results.

\begin{table*}[t]
\centering
\small
\begin{tabular}{lcc}
\toprule
Method & Mean of rater-level means & Mean of rater-level std. \\
\midrule
PREFINE (EPER)   & 7.82 & 0.98 \\
Self-Refine     & 6.70 & 1.11 \\
Prompt-Persona  & 5.39 & 1.53 \\
\bottomrule
\end{tabular}
\caption{
Rater-level score statistics in human evaluation.
For each annotator, we compute the mean and standard deviation of the scores assigned to each method across their evaluation sets,
and then aggregate these statistics across annotators.
}
\label{tab:rater-level-stats}
\end{table*}

\section{Additional Results with Mistral-7B}
\label{appendix:mistral-results}
To examine whether the effectiveness of PREFINE depends on the choice of the backbone model,
we conducted additional experiments using \texttt{Mistral-7B} as the backbone LLM.
In this setting, all components of PREFINE—including the generation, expert, pseudo-user, and refine agents—are instantiated using the same \texttt{Mistral-7B} model,
with differences between agent roles implemented solely through prompt design, following the same configuration as in the main experiments.

As in the main experiments, we kept the prompt design, refinement procedure, and evaluation protocol identical,
thereby isolating the effect of the backbone model choice.
Automatic evaluation was performed on both the PerDOC and PerMPST datasets
using the same LLM-based evaluation framework as in the main experiments.

Table~\ref{tab:comparison_winrates_mistral} reports the results on the PerDOC dataset.
The full configuration of PREFINE (EPER) substantially outperforms all baseline methods
and achieves stronger personalization performance than the model variants.

Table~\ref{tab:mpst-results-mistral} presents the results on the PerMPST dataset.
The Bayesian ordinal regression analysis shows that, for EPER,
the 95\% credible intervals of the latent score differences relative to other methods
lie entirely in the positive range, indicating consistently favorable performance.

Overall, across both PerDOC and PerMPST,
these results demonstrate that PREFINE (EPER) retains its effectiveness
when instantiated with \texttt{Mistral-7B},
confirming that the proposed framework generalizes beyond a specific backbone model.

\begin{table}[t]
\small
\centering
\begin{tabularx}{\columnwidth}{l|XXXXXXX}
\toprule
\diagbox{A}{B} & ZP & PP & SR & IPIR & EPIR & IPER & EPER \\
\midrule
\midrule
ZP          & -- & \cellcolor{myblue!74} 0.26 & \cellcolor{myblue!100} 0.00 & \cellcolor{myblue!100} 0.00 & \cellcolor{myblue!100} 0.00 & \cellcolor{myblue!100} 0.00 & \cellcolor{myblue!100} 0.00 \\
PP          & \cellcolor{myred!74} 0.74 & -- & \cellcolor{myblue!98} 0.02 & \cellcolor{myblue!99} 0.01 & \cellcolor{myblue!100} 0.00 & \cellcolor{myblue!100} 0.00 & \cellcolor{myblue!99} 0.01 \\
SR          & \cellcolor{myred!100} 1.00 & \cellcolor{myred!98} 0.98 & -- & \cellcolor{myblue!61} 0.39 & \cellcolor{myblue!87} 0.13 & \cellcolor{myblue!64} 0.36 & \cellcolor{myblue!84} 0.16 \\
\midrule
IPIR        & \cellcolor{myred!100} 1.00 & \cellcolor{myred!99} 0.99 & \cellcolor{myred!61} 0.61 & -- & \cellcolor{myblue!82} 0.18 & \cellcolor{myblue!51} 0.49 & \cellcolor{myblue!75} 0.25 \\
EPIR        & \cellcolor{myred!100} 1.00 & \cellcolor{myred!100} 1.00 & \cellcolor{myred!87} 0.87 & \cellcolor{myred!82} 0.82 & -- & \cellcolor{myred!74} 0.74 & \cellcolor{myblue!66} 0.34 \\
IPER        & \cellcolor{myred!100} 1.00 & \cellcolor{myred!100} 1.00 & \cellcolor{myred!64} 0.64 & \cellcolor{myred!51} 0.51 & \cellcolor{myblue!74} 0.26 & -- & \cellcolor{myblue!79} 0.21 \\
\midrule
\textbf{EPER} & \cellcolor{myred!100} 1.00 & \cellcolor{myred!99} 0.99 & \cellcolor{myred!84} 0.84 & \cellcolor{myred!75} 0.75 & \cellcolor{myred!66} 0.66 & \cellcolor{myred!79} 0.79 & -- \\
\bottomrule
\end{tabularx}
\caption{Win rate of the A-side model against the B-side model, averaged over five perspectives.
Stories were generated using Mistral-7B. Red/blue indicates that the row/column model is preferred.}
\label{tab:comparison_winrates_mistral}
\end{table}

\begin{table}[t]
\centering
\footnotesize
\setlength{\tabcolsep}{4pt} 
\begin{tabular}{lccccc}
\toprule
Method              & Score & $\Delta$ & 95\% CI & $P(\Delta>0)$ \\
\midrule
ZP              &   7.69±1.50     & 1.33 & [1.12, 1.53] &   1.00  \\
PP              &   7.83±1.49     & 0.58 & [0.34, 0.84] &    1.00  \\
SR              &    8.14±1.23    & 0.29 & [0.10, 0.48] &    1.00   \\
\midrule
IPIR                &  7.88±1.37    & 0.94 & [0.74, 1.14] &   1.00  \\
EPIR                &  8.02±1.23      & 0.73 & [0.53, 0.92] &  1.00   \\
IPER                &  8.01±1.26      & 0.69 & [0.48, 0.87] &   1.00  \\
\midrule
EPER & 8.23±1.18  & --  & -- & --         \\
\bottomrule
\end{tabular}
\caption{
Automatic evaluation results on the PerMPST dataset.
“Score” denotes the mean $\pm$ standard deviation of 10-point Likert ratings assigned by the LLM evaluator.
All stories were generated by Mistral-7B.
The table also reports results from a Bayesian ordinal regression model that treats Likert ratings as ordinal data. We report the posterior mean, 95\% credible interval, and posterior probability of the latent score difference $\Delta = \alpha_{\mathrm{EPER}} - \alpha_{m}$,
where $\alpha_m$ denotes the method-specific latent location parameter and $m$ is the compared method (baseline or model variant).}
\label{tab:mpst-results-mistral}
\end{table}

\section{Generated Plot Length Distribution}
\label{appendix:token_dist}
\begin{figure}[htbp]
  \centering
  \includegraphics[width=\columnwidth]{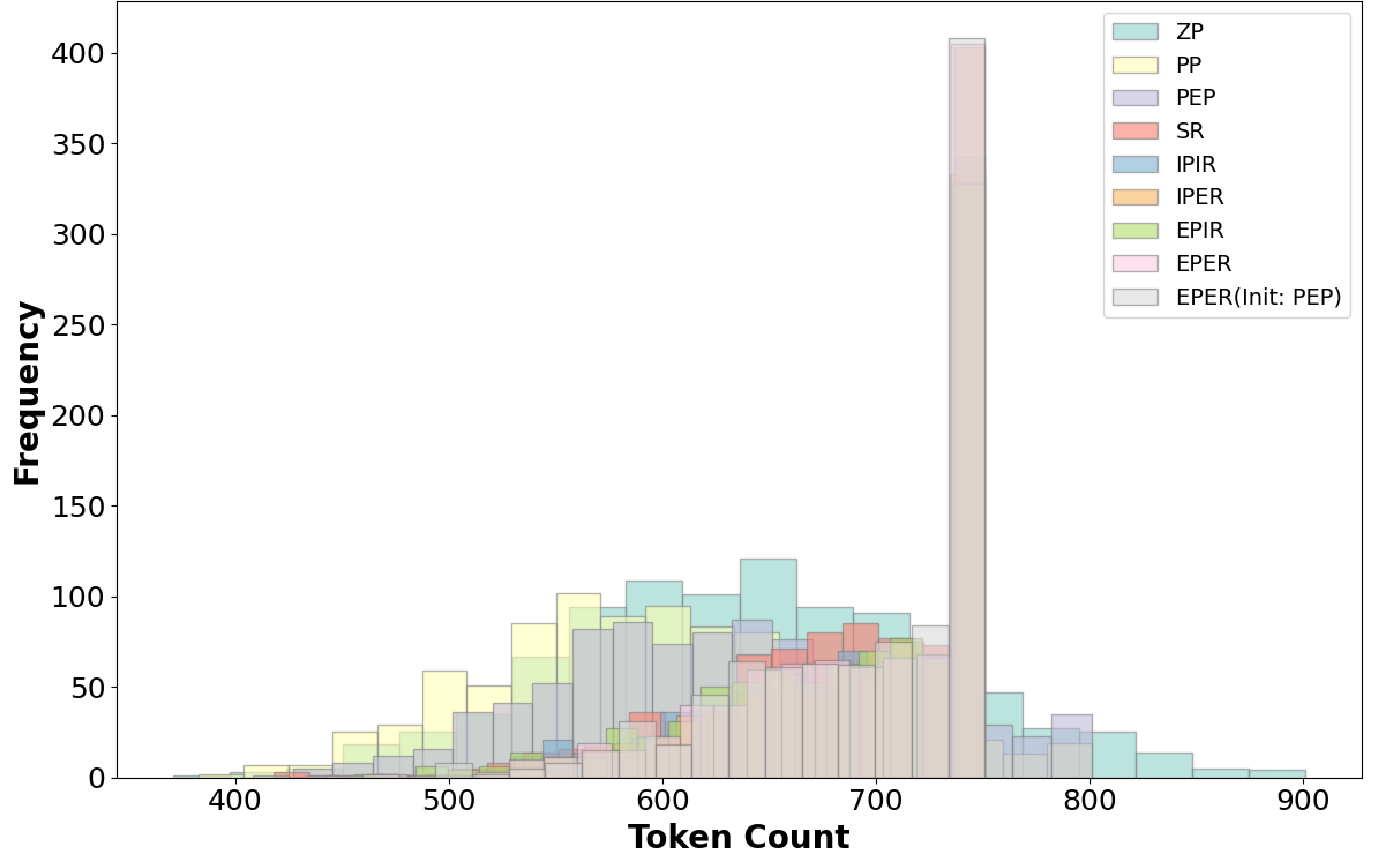}
  \caption{Distribution of generated story lengths (in tokens) in the PerDOC setting. 
Token lengths are measured using the LLaMA 3 tokenizer. 
Each method exhibits a different distribution, with some showing sharp peaks due to explicit length constraints in prompts.}
  \label{fig:tokencuntperdoc}
\end{figure}

\begin{figure}[htbp]
  \centering
  \includegraphics[width=\columnwidth]{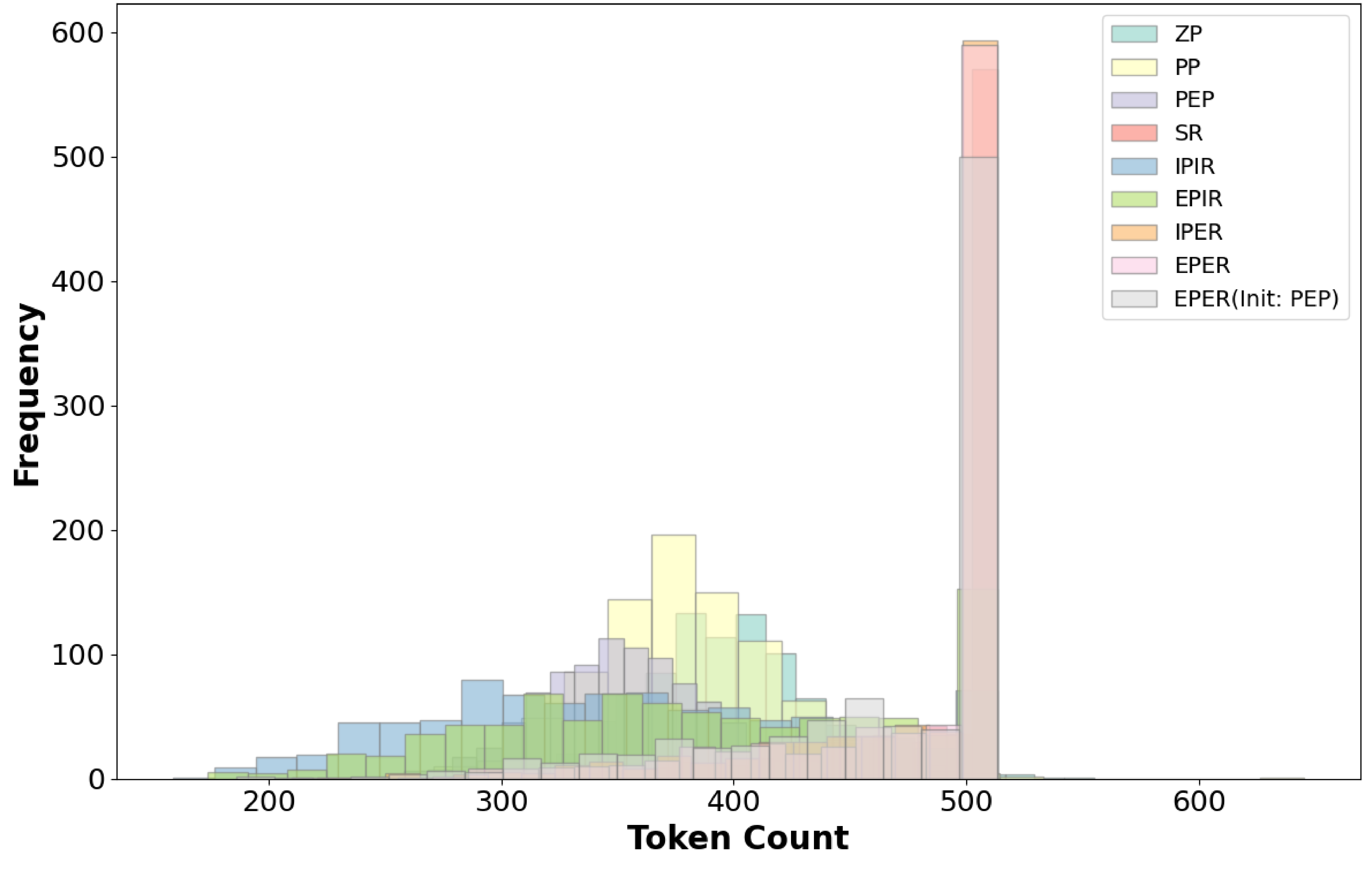}
  \caption{
Distribution of generated story lengths (in tokens) in the PerMPST setting. 
Token lengths are measured using the LLaMA 3 tokenizer. 
Each method exhibits a different distribution, with some showing sharp peaks due to explicit length constraints in prompts.
Compared to PerDOC, the generated stories tend to be shorter. 
}
  \label{fig:tokencntmpst}
\end{figure}

Figure~\ref{fig:tokencuntperdoc} shows the distribution of generated story lengths in the PerDOC setting, with the corresponding average and median values summarized in Table~\ref{tab:token_stats_perdoc}. Similarly, Figure~\ref{fig:tokencntmpst} and Table~\ref{tab:token_stats_permpst} present the distribution, average, and median of story lengths in the PerMPST setting.

As shown in Tables~\ref{tab:token_stats_perdoc} and~\ref{tab:token_stats_permpst}, stories generated in the PerDOC setting tend to be longer than those in the PerMPST setting.

The concentration of story lengths at specific token counts in some methods, as observed in Figures~\ref{fig:tokencuntperdoc} and~\ref{fig:tokencntmpst}, is due to explicit token-length constraints imposed via prompts. For methods that involve critique-and-refine loops (e.g., Self-Refine, EPER), we observed a tendency for story length to increase with each iteration if such constraints were not strictly enforced.

To ensure fair comparison across models and to accommodate the context length limitations of LLM judges, we applied explicit length constraints during generation. All token lengths reported here are measured using the LLaMA 3 tokenizer.
For example, Appendix~\ref{prompt:refinement-agent-doc}.

\begin{table}[htbp]
\centering
\small
\caption{Token length statistics of generated story plots (PerDOC). Mean and median token counts are shown for each method. Token lengths are measured using the LLaMA 3 tokenizer.}
\label{tab:token_stats_perdoc}
\begin{tabular}{lrrr}
\toprule
\textbf{Method} & \textbf{Mean} & \textbf{Median} \\
\midrule
Zero-Persona (ZP)              & 641.39 & 639.00  \\
Prompt–Persona (PP)         & 600.56 & 595.50 \\
Prompt–Expert-Persona (PEP)    & 631.51 & 628.00  \\
Self-Refine (SR)    & 690.12 & 703.00  \\
\midrule
IPIR                    & 692.03 & 709.00  \\
IPER                    & 701.17 & 721.00  \\
EPIR                    & 689.15 & 705.50  \\
EPER                    & 698.29 & 719.00  \\
EPER (Init: PEP)                   & 698.55 & 719.00  \\
\bottomrule
\end{tabular}
\end{table}

\begin{table}[htbp]
\centering
\small
\caption{Token length statistics of generated story plots (PerMPST). Mean and median token counts are shown for each method. Token lengths are measured using the LLaMA 3 tokenizer.}
\label{tab:token_stats_permpst}
\begin{tabular}{lrr}
\toprule
\textbf{Method} & \textbf{Mean} & \textbf{Median} \\
\midrule
Zero-Persona (ZP)              & 398.21 & 396.00 \\
Prompt–Persona (PP)         & 381.09 & 379.00 \\
Prompt–Expert-Persona (PEP)    & 352.11 & 352.00 \\
Self-Refine (SR)    & 486.52 & 513.00 \\
\midrule
IPIR                    & 356.88 & 350.00 \\
EPIR                    & 390.15 & 385.00 \\
IPER                    & 482.26 & 513.00 \\
EPER                    & 480.87 & 513.00 \\
EPER (Init: PEP)           & 467.72 & 512.00 \\
\bottomrule
\end{tabular}
\end{table}

\section{Additional Analysis}

\subsection{Relationship Between Token Length and LLM-Judge Evaluation on PerDOC}
To examine whether output length influenced model evaluation, we analyzed the relationship between token length and win rate on the PerDOC dataset.

\label{appendix:rel-perdoc-length} presents a subset analysis restricted to pairs whose token-length difference is within $\pm10$. EPER’s win rate changes little and remains superior to the baselines. As a result, the effect of length differences on win rate/score is quite limited.

\begin{figure}[htbp]
  \centering
  \includegraphics[width=\columnwidth]{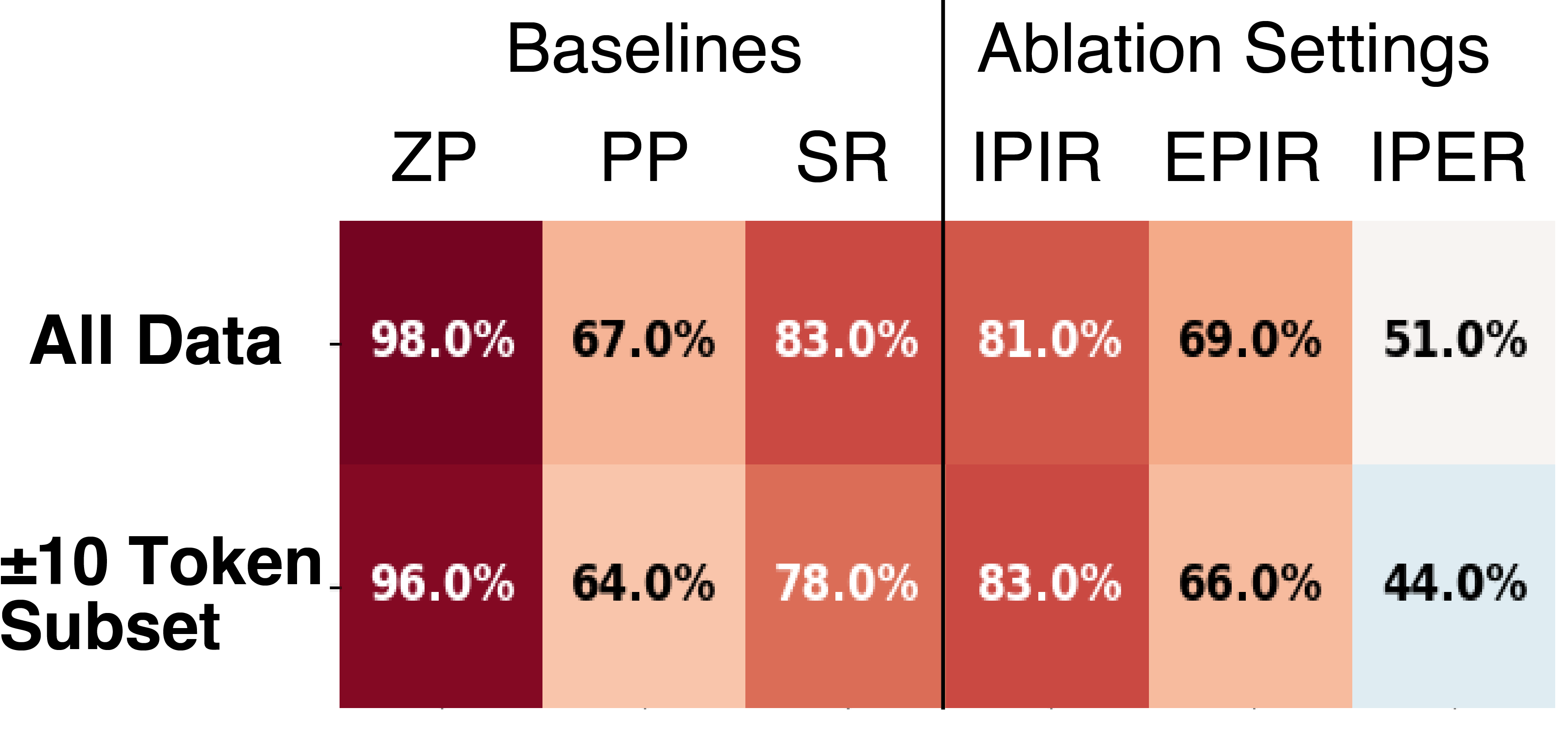}
  \caption{Relationship between token length and win rate for EPER outputs on PerDOC.}
  \label{fig:rel-token-length-perdoc}
\end{figure}

\subsection{Aspect-Wise Win Rate Analysis on PerDOC}
\label{appendix:aspect-winrate-perdoc}
In the PerDOC setting, the goal is to generate stories that align with user preferences on specific aspects. There are five evaluation aspects defined in the dataset: Interestingness, Surprise, Adaptability, Character Quality, and Ending Satisfaction \cite{perdoc}. Users’ interaction histories were also collected in alignment with these aspects.

\begin{figure*}[htbp]
  \centering
  \includegraphics[width=\textwidth]{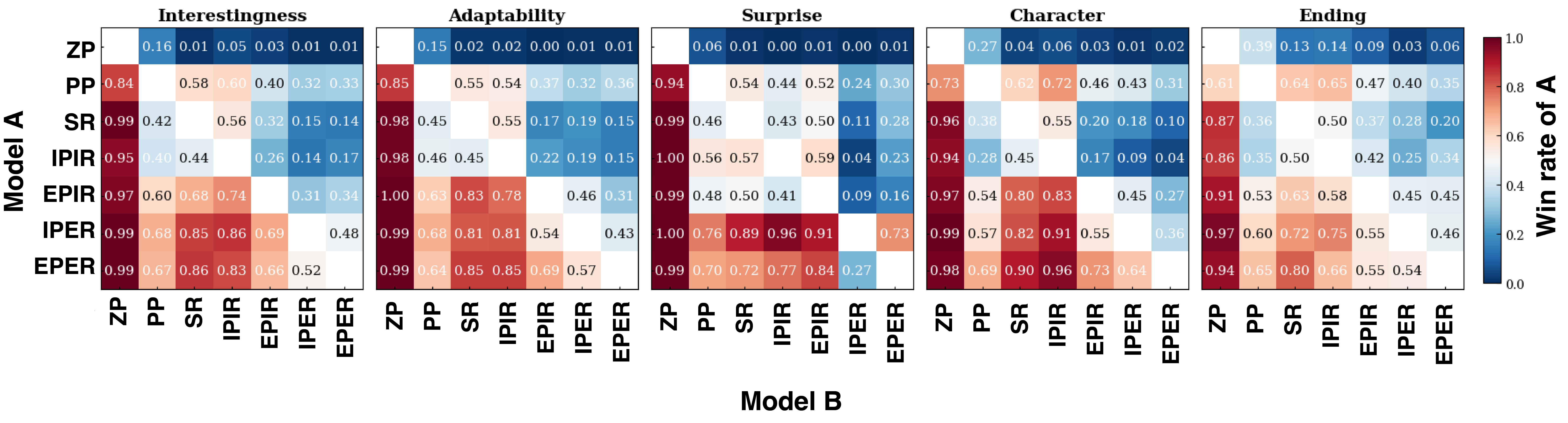}
  \caption{Aspect-wise win rate comparison in the PerDOC setting. 
Each matrix represents the win rate of the model on the A-side over the model on the B-side for each aspect.
EPER outperforms both baseline methods and its own variants across most aspects.
}
  \label{fig:appendix-winrate-aspect}
\end{figure*}

Figure~\ref{fig:appendix-winrate-aspect} shows the aspect-wise win rates for each method. As shown, our full method EPER (a complete configuration of PREFINE) achieves consistently higher win rates across most aspects compared to both baselines and its own variants. 

\subsection{Score-derived win rate on PerMPST}
\label{appendix:mpst-winrate}
In the automatic evaluation on PerMPST, we use an LLM-as-a-judge framework to assign 10-point scores (1–10) to each generated story.
In this section, we define a \emph{score-derived} win rate and analyze results from a pairwise perspective.

For a story pair $i$ generated by EPER and a comparator method $Y$, let $score_i^{\mathrm{EPER}}$ and $score_i^{Y}$ denote their scores.
We define a \emph{win} if $score_i^{\mathrm{EPER}}>score_i^{Y}$, a \emph{loss} if $score_i^{\mathrm{EPER}}<score_i^{Y}$, and a \emph{tie} if $score_i^{\mathrm{EPER}}=score_i^{Y}$ (scores are integers in $[1,10]$).
Excluding ties from the denominator, the win rate is
\[
\hat w \;=\; \frac{\#\text{wins}}{\#\text{wins}+\#\text{losses}}
\]
(\textit{ties excluded}).

Table~\ref{tab:mpst-winrate-results} reports the computed results.

\begin{table}[htbp]
\centering
\resizebox{0.95\linewidth}{!}{%
\begin{tabular}{lccc}
\toprule
Method              & EPER's Win rate  & $p$-value \\
\midrule
Zero-Persona              &   0.69    & $< 10^{-13}$        \\
Prompt-Persona          &     0.67   & $< 10^{-11}$      \\
Self-Refine         &     0.64   & $< 10^{-6}$        \\
\midrule
IPIR                &     0.47   & 0.279       \\
EPIR                &     0.50   & 1.00      \\
IPER                &   0.58    & 0.005            \\
\midrule
EPER & - & -           \\
\bottomrule
\end{tabular}
}
\caption{Score-derived win rates of EPER against each method on PerMPST (ties excluded). $p$-values from two-sided binomial tests}
\label{tab:mpst-winrate-results}
\end{table}

\subsection{Aspect-Wise Win Rate Analysis Starting from Prompt-Expert-Persona(PEP) Outputs on PerDOC}
\label{appendix:aspect-winrate-perdoc-pep}

\begin{figure*}[htbp]
  \centering
  \includegraphics[width=\textwidth]{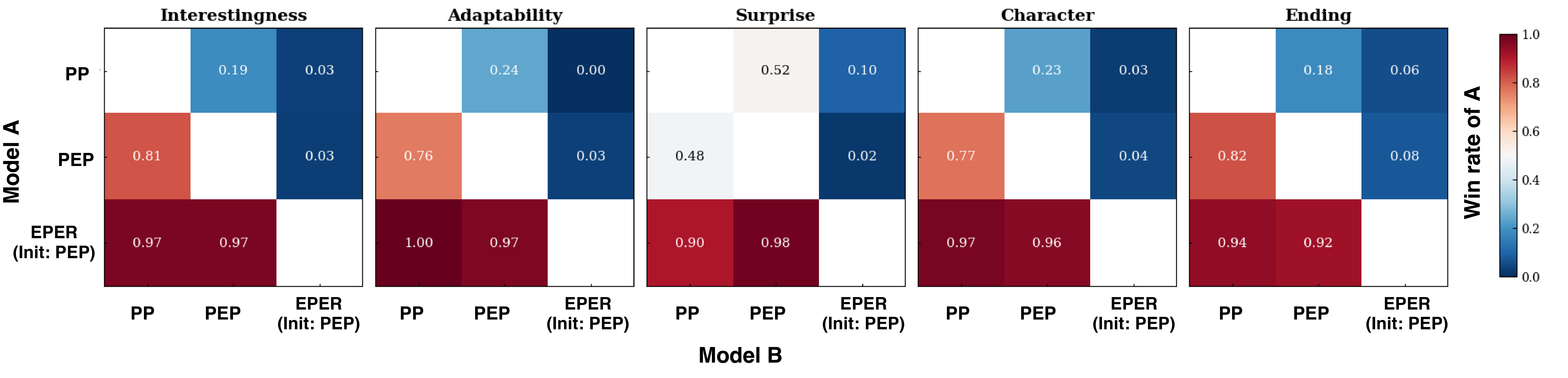}
  \caption{Aspect-wise win rate comparison among Prompt-Persona (PP), Prompt-Expert-Persona (PEP), and EPER (initialized with PEP outputs) in the PerDOC setting. 
PEP consistently outperforms PP across all aspects, demonstrating stronger personalization.
EPER further improves upon PEP, achieving the highest win rates across all aspects.
  }
  \label{fig:winrate-aspect2}
\end{figure*}

Figure~\ref{fig:winrate-aspect2} shows the aspect-wise win rates when EPER is applied to stories already personalized using Prompt-Expert-Persona (PEP). 
The results also include Prompt-Persona (PP) for reference. 
We observe that PEP consistently outperforms PP across all aspects, indicating that PEP already produces significantly personalized outputs.
Moreover, EPER further improves upon PEP in every aspect, demonstrating that our method is capable of enhancing personalization even when starting from an already strongly personalized story.

\subsection{Score Analysis Starting from Prompt-Expert-Persona(PEP) Outputs on PerMPST}
\label{appendix:mpst-pep}
Table~\ref{tab:mpst-results-pep} shows the results of applying EPER to stories already personalized using Prompt-Expert-Persona (PEP) in the PerMPST setting. 
We did not observe any significant improvement in scores when applying EPER over PEP. 
We attribute this to the limited opportunity for further personalization, possibly due to the relatively short length of the generated stories in the PerMPST style (see Appendix~\ref{appendix:token_dist}).

\begin{table}[htbp]
\centering
\caption{Comparison of scores between PEP and EPER (initialized from PEP outputs) on the PerMPST dataset,
along with results of two-sided Wilcoxon signed-rank tests ($n = 900$) comparing each method against EPER (Init: PEP).
No significant improvement was observed when applying EPER, suggesting limited room for further personalization.
.}
\label{tab:mpst-results-pep}
\resizebox{0.95\linewidth}{!}{%
\begin{tabular}{lccc}
\toprule
Method              & Score (mean ± std) & $p$-value \\
\midrule
Prompt-Persona (PP)          & 7.23 ± 1.47  &   $< 10^{-10}$   \\
Prompt-Expert-Persona (PEP) & 7.45±1.39     &    0.96    \\
EPER (Init: PEP)            &  7.45 ± 1.38  & -      \\
\bottomrule
\end{tabular}
}
\end{table}

\subsection{Analysis of Potential Evaluation Bias in User-Specific Rubric Suitability Rating}
\label{appendix:rubric_bias_analysis}
In the main paper (Section~\ref{User-Specific Rubric Quality}), we observed that users who rated the generated rubric as well-aligned with their preferences tended to assign higher scores to the stories generated by PREFINE(EPER). However, this effect may be partially influenced by individual evaluation biases, such as a general tendency to give higher scores.

\begin{figure}[htbp]
  \centering
  \includegraphics[width=\columnwidth]{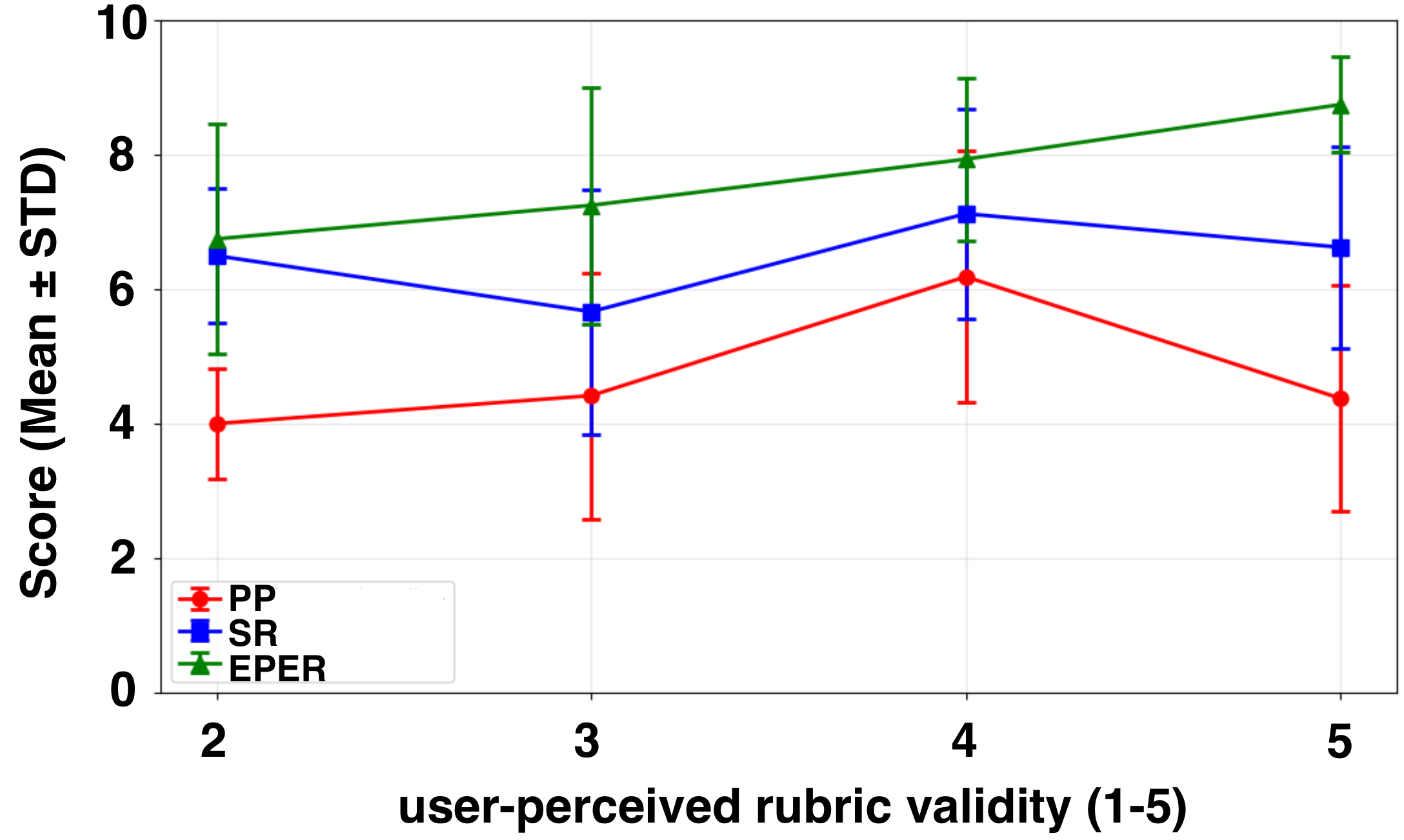}
  \caption{The relationship between users’ ratings of rubric suitability (on a 1–5 scale) and their average story scores for each generation method (PP, SR, EPER). While no clear trend is observed for PP and SR, EPER shows a visible correlation: users who perceived the rubric as more aligned with their preferences tended to assign higher scores to the generated stories.
}
  \label{fig:appendix-rubric-bis}
\end{figure}

To investigate this possibility, we analyzed whether participants who gave high rubric ratings also consistently assigned higher scores to stories produced by other methods (e.g., Prompt-Persona (PP), Self-Refine (SR)). The results are shown in Figure~\ref{fig:appendix-rubric-bis}.

Figure~\ref{fig:appendix-rubric-bis} shows no clear relationship between rubric–suitability ratings and story scores for PP or SR. In contrast, EPER exhibits a clear trend: users who rate the rubric highly also assign higher story scores.

This suggests that perceived rubric validity may be meaningfully related to the effectiveness of personalization in PREFINE.

\subsection{Aspect-Wise General Story Quality Scores}
\label{appendix:general-rubric-quality}

\begin{figure}[h]
  \centering
  \includegraphics[width=\columnwidth]{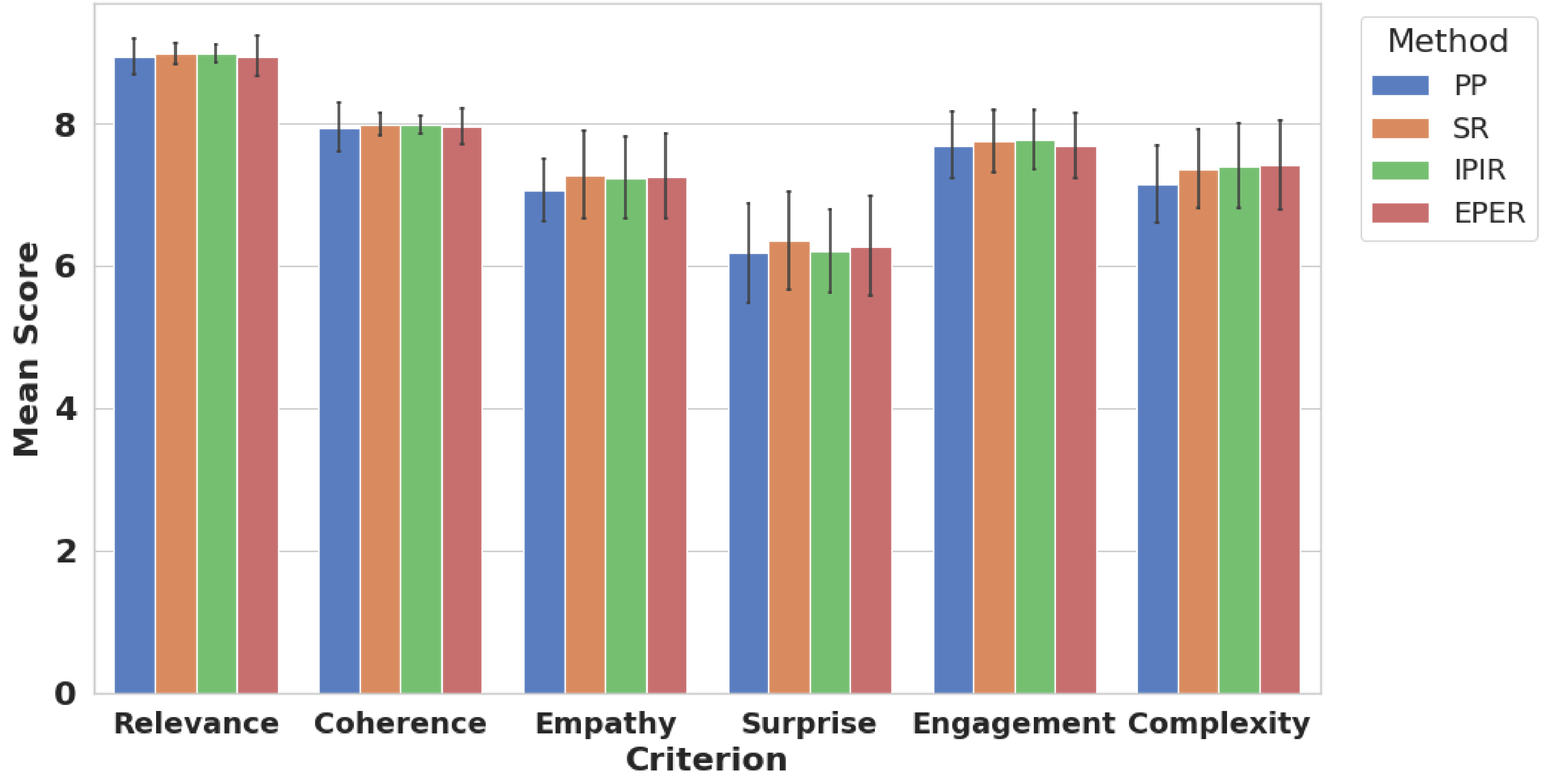}
  \caption{Aspect-wise story quality scores based on the six criteria defined in \cite{hanna}, as evaluated by GPT-4o on random subsets (200 stories each) from PerDOC and PerMPST. Error bars indicate standard deviation across samples. EPER achieves comparable quality to Self-Refine (SR) across most aspects, despite being optimized for personalization.}
  \label{fig:gr}
\end{figure}

Figure~\ref{fig:gr} presents the aspect-wise scores for story quality, as evaluated by the GPT-4o judge based on the six criteria defined in the HANNA framework \cite{hanna}. The evaluation was conducted on two random subsets of 200 stories each, sampled from the PerDOC and PerMPST outputs respectively.

Across most aspects, EPER achieves scores comparable to Self-Refine (SR), a method explicitly designed to enhance story quality according to general rubrics. This is particularly notable, as EPER was not directly optimized for these criteria but instead for user personalization. These results suggest that EPER is capable of maintaining strong general story quality—on par with established baselines—while simultaneously achieving effective personalization.

\section{Implementation and Reproducibility Details}
\label{appendix:Implementation}
All experiments were conducted with a fixed random seed (seed = 42) to ensure consistency across runs. Due to inference cost constraints, each configuration was executed once. The temperature was set to 0.7 for story generation tasks and 0 for evaluation.

\paragraph{Model Configuration.}
We used \texttt{LLaMA-3-70B} for some agent, accessed via the \texttt{Together.ai}\footnote{https://www.together.ai/} API. Fine-tuning models (e.g., PerSE-Llama3(8B)) was also performed on the Together.ai platform, following the same training settings as in \citet{perse}, including dataset splits, learning rates, and number of training steps, etc. 

\paragraph{Computing Environment.}
All other procedures, including data preprocessing, model inference using \texttt{PerSE-Llama3-8B} and \texttt{Mistral-7B}, and result analysis, were conducted on a local machine running Ubuntu 24.04.2 LTS with an Intel Xeon w5-2465X CPU, 256 GB of RAM, and two NVIDIA RTX 6000 Ada GPUs (48 GB each). The software environment included Python 3.11.4 and CUDA 12.9. We used PyTorch and the HuggingFace Transformers library for model implementation and inference.

\subsection{Personalized Refinement Example from \textsc{PerMPST}}
We present a concrete example illustrating how \textsc{PREFINE} performs personalized story refinement using a user-specific rubric under the \textit{EPER} configuration.

\paragraph{User-Specific Rubric}

The following rubric was generated to reflect a particular user's preferences, based on their interaction history:
\begin{itemize}
    \item The story features complex, high-stakes situations that drive the plot forward.
    \item Characters are authentic, multi-dimensional, and willing to challenge authority.
    \item The narrative incorporates realistic and innovative storytelling techniques.
    \item The plot explores complex moral issues and emotional themes in a nuanced manner.
    \item The story has a strong sense of autonomy and self-determination in its characters' actions.
\end{itemize}

Using this rubric, the critique-and-refine process was iteratively conducted. Table~\ref{tab:refine-cycle} illustrates an example refinement cycle, including the initial story, generated critiques, and corresponding improved versions.

\begin{table*}[ht]
\centering
\scriptsize 
\caption{Example of a critique-and-refine cycle for personalization using \textsc{PREFINE} under the EPER configuration.
This example illustrates the process from the initial plot generation through the second round of critique. It demonstrates how the output is progressively improved according to the user-specific rubric. While the full refinement consists of seven iterative cycles, only the first part is shown here due to space limitations.}
\label{tab:refine-cycle}
\resizebox{0.8\textwidth}{!}{%
\begin{tabularx}{\textwidth}{|p{2.5cm}|>{\raggedright\arraybackslash}X|}
\hline
\textbf{Stage} & \textbf{Content} \\
\hline
Initial Story $s^{(0)}$& In 1951, Zane Carpenter and Trace Allen are high-school seniors and friends in Willow Creek, a small, declining Southern United States town. As they navigate the complexities of adolescence, they find themselves at a crossroads, torn between pursuing their individual dreams and staying in their hometown to support their families. Zane, the charismatic and confident one, has always been destined for greatness, with a scholarship to a prestigious university awaiting him. Trace, on the other hand, is more reserved and introspective, with a passion for writing that he hopes to turn into a career. Despite their differences, they share a deep bond, forged through years of shared experiences and adventures in the small town. One fateful night, they stumble upon a mysterious and beautiful stranger, Emily, who has just arrived in Willow Creek, and their lives are forever changed. Emily's presence awakens a sense of restlessness and longing in both Zane and Trace, forcing them to confront their own desires and aspirations. As they spend more time with Emily, they begin to see their town and their lives in a different light, and the possibility of escape and new beginnings becomes tantalizingly real. The trio's friendship deepens, but it also creates tension and conflict, particularly when Zane and Trace find themselves falling for Emily's charms. As the summer of 1951 draws to a close, Zane, Trace, and Emily must make difficult choices that will determine the course of their lives and the future of their relationships. The town of Willow Creek, with its faded grandeur and narrow expectations, serves as a poignant backdrop for their struggles and triumphs. Ultimately, the friends must decide whether to stay in the only home they have ever known or to take a chance on the unknown, and the consequences of their decisions will haunt them for years to come. With the fate of their friendships and their futures hanging in the balance, Zane, Trace, and Emily must navigate the complexities of love, loyalty, and ambition in a small town where everyone knows each other's secrets. \\
\hline
Feedback $F^{(1)}$  & Criterion: The story features complex, high-stakes situations that drive the plot forward.

Score: 6

Explanation: The plot revolves around the characters' personal struggles and relationships, but the stakes could be higher.

Suggestion: Introduce external conflicts, such as economic or social pressures, to raise the stakes and create more tension.

Criterion: Characters are authentic, multi-dimensional, and willing to challenge authority.

Score: 7

Explanation: Zane and Trace have distinct personalities, but their willingness to challenge authority is not fully explored.

Suggestion: Show the characters questioning or defying the town's expectations and norms to add depth to their personalities.

Criterion: The narrative incorporates realistic and innovative storytelling techniques.

Score: 5

Explanation: The narrative is straightforward and lacks unique storytelling elements.

Suggestion: Incorporate non-linear storytelling or unconventional narrative structures to add innovation.

Criterion: The plot explores complex moral issues and emotional themes in a nuanced manner.

Score: 8

Explanation: The story touches on themes of love, loyalty, and ambition, but could delve deeper into moral complexities.

Suggestion: Introduce gray areas and conflicting values to create more nuanced moral dilemmas.

Criterion: The story has a strong sense of autonomy and self-determination in its characters' actions.

Score: 7

Explanation: The characters' decisions are influenced by their relationships and the town's expectations.

Suggestion: Emphasize the characters' independent decision-making and self-directed actions to increase their autonomy. \\
\hline
Refined Story $s^{(01)}$ & In 1951, Zane Carpenter and Trace Allen are high-school seniors and friends in Willow Creek, a small, declining Southern United States town. As they navigate the complexities of adolescence, they find themselves at a crossroads, torn between pursuing their individual dreams and staying in their hometown to support their families amidst the looming threat of the town's textile mill closure, which would devastate the local economy. Zane, the charismatic and confident one, has always been destined for greatness, with a scholarship to a prestigious university awaiting him, but he begins to question the town's expectations and the true cost of his ambition. Trace, on the other hand, is more reserved and introspective, with a passion for writing that he hopes to turn into a career, and he starts to secretly write for an underground newspaper that challenges the town's conservative values. Their lives are forever changed when they stumble upon a mysterious and beautiful stranger, Emily, who has just arrived in Willow Creek, and they find themselves drawn to her free-spirited nature and her own moral dilemmas, as she is hiding a dark secret that could ruin her reputation in the town. As the summer of 1951 unfolds in a non-linear narrative, the trio's friendship deepens, but it also creates tension and conflict, particularly when Zane and Trace find themselves falling for Emily's charms, and they must confront the gray areas between love, loyalty, and ambition. Through a series of flashbacks and fragmented memories, the story reveals the characters' inner struggles and their willingness to challenge authority, as they make difficult choices that will determine the course of their lives and the future of their relationships, ultimately deciding whether to stay in the only home they have ever known or to take a chance on the unknown, with their autonomy and self-determination guiding their decisions.\\
\hline
Feedback $F^{(2)}$ & Criterion: The story features complex, high-stakes situations that drive the plot forward.

Score: 7

Explanation: The plot has a clear conflict with the textile mill closure, but it's somewhat predictable.

Suggestion: Introduce unexpected twists, like a personal tragedy or a town-wide crisis.

Criterion: Characters are authentic, multi-dimensional, and willing to challenge authority.

Score: 8

Explanation: Zane and Trace have distinct personalities, but Emily's character feels somewhat mysterious.

Suggestion: Add more depth to Emily's backstory and motivations.

Criterion: The narrative incorporates realistic and innovative storytelling techniques.

Score: 8

Explanation: The non-linear narrative is engaging, but the use of flashbacks is not particularly innovative.

Suggestion: Experiment with unconventional narrative structures or formats.

Criterion: The plot explores complex moral issues and emotional themes in a nuanced manner.

Score: 9

Explanation: The story tackles themes of ambition, loyalty, and identity in a thoughtful way.

Suggestion: Delve deeper into the gray areas between right and wrong.

Criterion: The story has a strong sense of autonomy and self-determination in its characters' actions.

Score: 8

Explanation: The characters make difficult choices, but their decisions feel somewhat influenced by external factors.

Suggestion: Emphasize the characters' inner drives and desires to make their choices feel more self-directed. \\
\hline
\end{tabularx}
}
\end{table*}

\end{document}